\documentclass{article}

 \usepackage[preprint]{neurips_2026}

\usepackage[utf8]{inputenc} 
\usepackage[T1]{fontenc}    
\usepackage{hyperref}       
\usepackage{url}            
\usepackage{booktabs}       
\usepackage{amsfonts}       
\usepackage{nicefrac}       
\usepackage{microtype}      
\usepackage{xcolor}         
\usepackage{graphicx}
\usepackage{amsmath}
\usepackage{float} 
\usepackage{amssymb}

\usepackage{adjustbox}
\usepackage{graphicx} 
\usepackage{pifont}  
\usepackage{enumitem}
\usepackage{tabularx}
\usepackage{booktabs}

\usepackage{amsmath}

\usepackage{multirow}
\usepackage{xcolor}
\usepackage[dvipsnames]{xcolor}

\usepackage[utf8]{inputenc} 
\usepackage[T1]{fontenc}    
\usepackage{hyperref}       
\usepackage{url}            
\usepackage{booktabs}       
\usepackage{amsfonts}       
\usepackage{nicefrac}       
\usepackage{microtype}      
\usepackage{xcolor}         

\title{Pretraining Induces a Reusable Spectral Basis for Downstream Task Adaptation

}

\author{%
Junjie Yu$^{1}$ \thanks{J. Yu, Y. Wang and Z. Deng contributed equally.} \quad Yue Wang$^{1, 2 \ *}$ \quad Zihan Deng$^{1 \ *}$ \quad Yan Zhu$^1$ \quad Wenxiao Ma$^1$  \quad \textbf{Quanying Liu}$^1$ \thanks{Corresponding author.} \\
$^1$Department of Biomedical Engineering, Southern University of Science and Technology \\ $^2$Department of Psychology, The University of Hong Kong\\
\texttt{\{12231192, 12112610, 12211804, 12212557, 12212659\}@mail.sustech.edu.cn}\\
\texttt{liuqy@sustech.edu.cn}
}

\begin{document}

\maketitle

\begin{abstract}
Finetuning pretrained models occurs in a low-dimensional subspace of the full parameter space. Prior work has focused on characterizing this optimization subspace, but largely ignored the complementary question: why do certain directions remain unexplored during finetuning? Are these stable directions irrelevant to downstream tasks, or do they already encode task-relevant structure that requires no further adjustment? Answering this question is central to understanding how pretrained knowledge transfers. Through systematic spectral analysis across vision and language models, we show that the leading singular vectors of pretrained weight matrices remain highly stable under finetuning and are shared across unrelated downstream tasks, revealing that pretraining establishes a reusable spectral coordinate system. Models pretrained on larger datasets exhibit greater spectral stability under distribution shift or task change, directly linking pretraining scale to geometric transferability. Motivated by these findings, we propose a parameter-efficient method that freezes pretrained singular vectors and optimizes only leading spectral coefficients, achieving competitive performance on GLUE with 0.2\% trainable parameters. Our results reveal that the stable directions encode transferable structure rather than irrelevant noise: successful pretraining discovers spectral bases that downstream tasks inherit and operate within.
\end{abstract}





\section{Introduction}

\begin{figure*}[!t]
  \centering
  \includegraphics[width=0.98\linewidth]{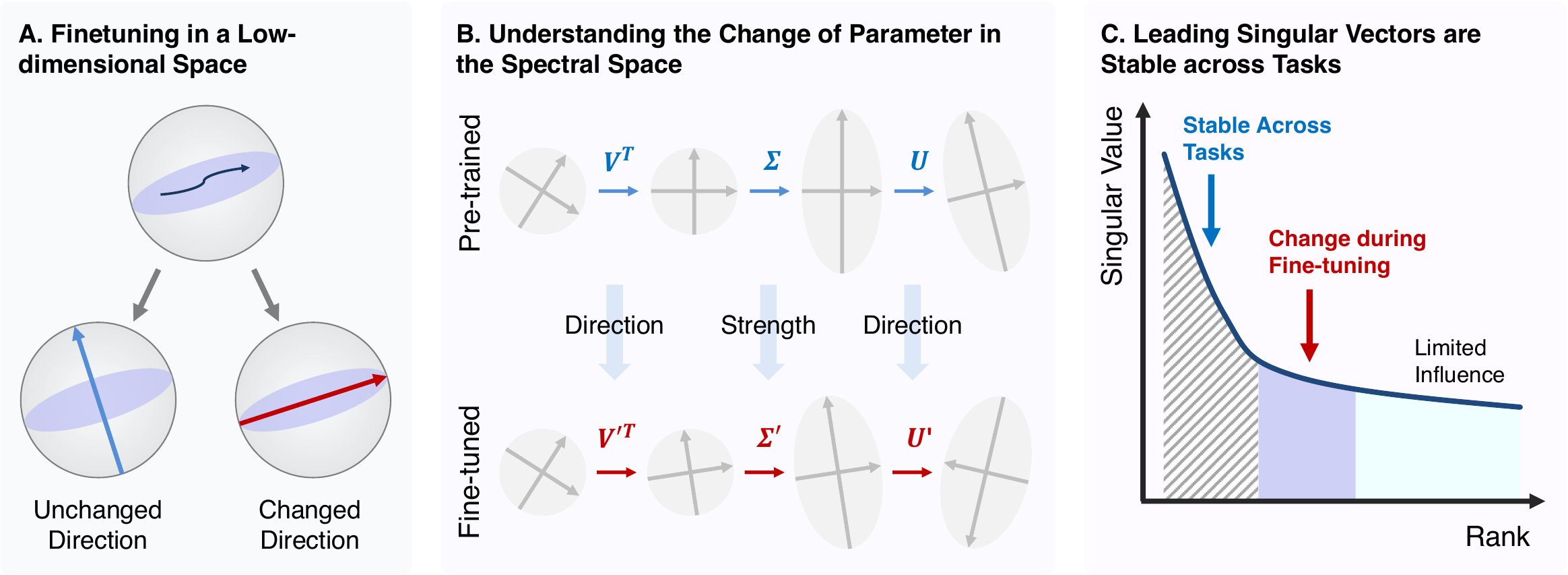}
  \caption{
  \textbf{Pretraining induces a reusable spectral basis for downstream task adaptation.} 
    \textbf{(A)} Finetuning updates are not uniformly distributed across the parameter space: some directions undergo significant changes while others remain largely unchanged, suggesting that optimization is effectively confined to a low-dimensional subspace.
    \textbf{(B)} To understand which directions are preserved and which are altered during finetuning, we decompose weight matrices via SVD and analyze the changes in singular values and singular vectors before and after finetuning.
    \textbf{(C)} The leading singular vectors remain nearly unchanged after finetuning and are consistently shared across different downstream tasks, revealing a task-agnostic spectral basis induced by pretraining.
  }
  \label{fig:motivation}
\end{figure*}

Finetuning a pretrained model on a downstream task is the dominant paradigm in machine learning \citep{wu2025llm, han2024parameter}. A well-established observation is that parameter changes during finetuning are geometrically concentrated: full finetuning, despite updating every weight, produces changes that lie in a much smaller intrinsic subspace than the full parameter space~\citep{aghajanyan2021intrinsic, zhang2023fine, fukuhata2025few}. This observation has motivated an extensive line of work on where finetuning optimizes, identifying the low-dimensional subspace that captures most of the task-relevant update~\citep{zhang2023fine, meng2024pissa}. Yet this focus on the directions that \emph{change} has left an equally important question unexamined: what about the directions that \emph{do not change} (Figure \ref{fig:motivation}A)?

The directions that finetuning leaves intact may be the more informative half of the story. When gradient descent consistently avoids rotating certain directions across diverse tasks and architectures, there are two possible explanations: either these directions are irrelevant to downstream performance and finetuning ignores them because updating them would not help, or they are already well-aligned with what downstream tasks require and are preserved precisely because they are useful \citep{neyshabur2020being, yosinski2014transferable, alain2016understanding}. These two explanations carry fundamentally different implications for how we understand pretrained model transferability. In the first picture, pretrained weights are merely a convenient initialization. In the second, they encode a coordinate system that downstream tasks inherit and operate within, and the low-dimensional structure of finetuning reflects this inherited geometry rather than an accidental byproduct of optimization. Characterizing the stable directions mathematically is therefore central to understanding why certain pretrained models transfer better than others and how to exploit this structure for more principled adaptation.

We investigate these questions through singular value decomposition (SVD), which provides a natural mathematical language for decomposing the directional structure of weight matrices. SVD factorizes each weight matrix into orthogonal directions (singular vectors) and their associated strengths (singular values), separating \emph{which} directions a layer is sensitive to from \emph{how strongly} it responds along each direction. This decomposition lets us ask not only whether finetuning rotates the singular vectors of pretrained weights or merely adjusts their singular values, but also, if rotation does occur, precisely \emph{which} directions are rotated and which are left intact (Figure \ref{fig:motivation}B). This is exactly the granularity needed to characterize the stable directions whose origin and mathematical structure we seek to understand.


Our analysis reveals a systematic and interpretable answer. The leading singular vectors, those associated with the large singular values, remain stable under finetuning across diverse tasks, architectures, and modalities, even when parameter updates are substantial and performance improves significantly (Figure \ref{fig:motivation}C). This stability is not merely a consequence of small gradient steps: by decomposing finetuning updates via SVD, we find that even when update magnitudes are large, their singular vectors remain strongly aligned with those of the pretrained weights, actively reinforcing existing directions rather than rotating them. This reveals a coupling between the spectral geometry of pretrained weights and the gradient directions that finetuning follows.

We further explore the formation of this coordinate system during pretraining. By analyzing checkpoints throughout pretraining, we find that leading singular vectors converge progressively: the similarity between top singular vectors across epochs increases steadily during training and reaches near-perfect alignment in later stages, indicating that pretraining discovers and stabilizes a consistent set of directions. We also show that models pretrained on larger datasets exhibit greater spectral stability when further trained on new tasks or data distributions. This reveals that larger pretraining scale yields spectral bases that are more transferable across diverse downstream contexts.



These findings reveal a clear spectral hierarchy in pretrained models. Leading singular vectors remain highly stable across tasks and distributions, indicating they encode reusable structure inherited from pretraining. Besides, trailing singular vectors have negligible singular values and contribute little to performance. This suggests that task-specific adaptation primarily occurs through reweighting of the dominant spectral components, those with sufficient magnitude to influence model behavior yet not so rigidly fixed as to resist adjustment.

To validate this hypothesis, we propose a minimal parameterization: optimize only the top-$K$ spectral coefficients, the upper-left $K \times K$ block of the singular value matrix, while freezing all singular vectors. This design directly tests whether selective reweighting within the pretrained spectral basis suffices for adaptation.

The finetuned weight is expressed as $W = U \Sigma' V^\top$, where $\Sigma'$ differs from the pretrained $\Sigma$ only in its top-left $K \times K$ block. Remarkably, when applied to DeBERTa-V3 across a range of language tasks, this approach achieves performance comparable to existing parameter-efficient finetuning methods while updating only 0.2\% of the parameters. This confirms that task adaptation can indeed be accomplished through selective reweighting within the pretrained spectral basis, validating the central role of dominant spectral components in transfer learning.

Our main contributions are as follows:

\begin{itemize}[leftmargin=*]
    \item We demonstrate that leading singular vectors of pretrained weight matrices remain stable during finetuning and are shared across diverse tasks, revealing a reusable spectral coordinate system inherited from pretraining.
    
    \item We show that larger pretraining data scale increases the transferability of leading singular vectors, making them more transferable under distribution shift and new tasks.
    
    \item Motivated by these findings, we propose a parameter-efficient finetuning method that freezes pretrained singular vectors and optimizes only the leading spectral coefficients, achieving competitive performance with substantially fewer trainable parameters.
\end{itemize}

\section{Related Work}





\paragraph{Parameter-efficient fine-tuning and low-dimensional adaptation.}
Full-parameter finetuning exhibits a striking low-dimensional structure: although all parameters are updated, the resulting changes lie in a much smaller intrinsic subspace than the full parameter space~\citep{aghajanyan2021intrinsic, zhang2023fine, li2018measuring, fukuhata2025few}. This observation has motivated extensive work on identifying which low-dimensional subspace captures task-relevant updates~\citep{zhang2023fine, meng2024pissa}. Parameter-efficient fine-tuning (PEFT) methods operationalize this view by showing that training only a small fraction of parameters can match full finetuning across varied tasks~\citep{houlsby2019parameter, hu2022lora, zaken2022bitfit, zhang2023adalora, liu2024dora, he2021towards}. 
However, prior work focuses primarily on the directions that \emph{change} during finetuning, leaving unexamined the equally important question of which directions remain \emph{stable} and why. Understanding whether stable directions reflect inherited structure from pretraining or merely irrelevant parameters is central to explaining pretrained model transferability~\citep{neyshabur2020being, yosinski2014transferable}. Our work addresses this gap by systematically characterizing the spectral stability of pretrained weights and tracing its origin to the pretraining process itself.

\paragraph{Spectral analysis of neural network parameters.}
Singular value decomposition provides a natural mathematical framework for understanding parameter changes by decomposing weight matrices into orthogonal directions (singular vectors) and their associated strengths (singular values). This decomposition enables us to characterize adaptation from two complementary perspectives: whether finetuning modifies the \emph{directions} a layer is sensitive to or merely adjusts \emph{how strongly} it responds along existing directions.
Prior spectral analyses have revealed important structural properties. Singular-value spectra of trained networks exhibit heavy-tailed distributions~\citep{martin2021implicit, mahoney2019traditional}, indicating that parameter information concentrates in a small number of leading directions. Singular-vector analyses are more challenging due to incompatible parameter dimensions across architectures. Prior work addresses this by analyzing representations rather than parameters: dimensionality reduction projects different models into a common space where singular-vector alignment reveals that different neural networks learn similar representations~\citep{raghu2017svcca}.
However, these spectral tools have not been systematically applied to analyze the parameter structure and finetuning process. Our work fills this gap by demonstrating that leading singular vectors form a reusable coordinate system established during pretraining and preserved throughout finetuning.

\paragraph{Pretraining scale and transferability.}
Scaling-law studies show that increasing model and data scale improves downstream generalization \citep{kaplan2020scaling, hoffmann2022training, muennighoff2023scaling, hernandez2021scaling}, but focus on performance metrics rather than mechanistic explanations. Our work provides a geometric perspective: larger pretraining corpora yield more transferable singular-vector directions, directly linking scale to the structural quality of learned weight representations.

\section{Preliminaries}

\subsection{Spectral Coordinates and Similarity Measures}

For a weight matrix $W \in \mathbb{R}^{m \times n}$, the singular value decomposition (SVD) is $W = U \Sigma V^\top$, where the columns of $U$ and $V$ are the left and right singular vectors, and $\Sigma$ is a diagonal matrix of non-negative singular values in decreasing order. The singular vectors define a natural directional coordinate system for the weight matrix, decoupling the directions of computation from their magnitudes.

To characterize how finetuning modifies each layer, we measure the alignment between pretrained and finetuned singular vectors via cosine similarity. Concretely, for the $i$-th singular vector pair, we compute $\cos\theta_i = |\tilde{u}_i^\top u_i|$ and $|	\tilde{v}_i^\top v_i|$, 

where $u_i, v_i$ and $\tilde{u}_i, \tilde{v}_i$ are the $i$-th left and right singular vectors of the pretrained and finetuned weight matrices, respectively. Values close to 1 indicate that the corresponding directions are preserved after finetuning, while values close to 0 indicate substantial rotation.
We use the absolute value because singular vectors are defined only up to a simultaneous sign flip: replacing $(u_i, v_i)$ with $(-u_i, -v_i)$ leaves the rank-1 component $\sigma_i u_i v_i^\top$, and hence the matrix $W$, unchanged. Therefore, inner products of $1$ and $-1$ both represent perfect alignment in the SVD basis.


\subsection{Analysis of Cumulative Parameter Update}
\label{sec:method_grad}

In Section~\ref{sec:gradient}, we further analyze the cumulative parameter update $\Delta W = W^{\text{ft}} - W^{\text{pt}}$, which represents the model change in weights after finetuning. Since each gradient step contributes an additive increment, $\Delta W$ is the accumulated sum of all gradient updates applied throughout training, and we therefore refer to it as the \emph{cumulative gradient}.

Specifically, we examine two key aspects. First, we compute the ratio of singular values between $\Delta W$ and $W^{\text{pt}}$ at corresponding ranks to quantify the relative magnitude of change:
\begin{equation}
r_i = \frac{\sigma_i(\Delta W)}{\sigma_i(W^{\text{pt}})},
\end{equation}
where $\sigma_i(\cdot)$ denotes the $i$-th singular value. This ratio assesses the extent of parameter change relative to the original pretrained parameters.




Second, we quantify directional alignment between the update $\Delta W$ and the pretrained weight $W^{\text{pt}}$ by measuring the cosine similarity of their singular vectors at corresponding ranks:
\begin{equation}
a_i^{(u)} = \left| \langle u_i(\Delta W), u_i(W^{\text{pt}}) \rangle \right|, \quad
a_i^{(v)} = \left| \langle v_i(\Delta W), v_i(W^{\text{pt}}) \rangle \right|,
\end{equation}
where $u_i(\cdot)$ and $v_i(\cdot)$ denote the $i$-th left and right singular vectors, respectively. The absolute value accounts for the sign ambiguity of singular vectors, and larger values indicate stronger alignment between the update and the pretrained spectral structure.





\subsection{Spectrally Restricted Finetuning}
\label{sec: SRF}

Our spectral analysis reveals that the leading singular vectors of pretrained weights remain highly stable after finetuning, while trailing components have negligible singular values. This motivates a simple adaptation strategy: freeze the pretrained singular vectors and optimize only within the subspace spanned by the top-$k$ directions.

Concretely, let $W^{\text{pt}} = U \Sigma V^\top$ be the pretrained SVD. We parameterize the finetuned weight as:

\begin{equation}
W = U (\Sigma + A) V^\top,
\end{equation}

where $A$ is constrained to be nonzero only in its leading $k \times k$ block:

\begin{equation}
A =
\begin{bmatrix}
M & 0 \\
0 & 0
\end{bmatrix},
\qquad
M \in \mathbb{R}^{k \times k}.
\end{equation}

The learnable matrix $M$ allows the model to reweight and linearly combine the top-$k$ pretrained spectral components without introducing new singular directions. This reduces trainable parameters from $m \times n$ to $k^2$ per weight matrix, while testing our hypothesis that the pretrained spectral basis suffices for downstream adaptation.

\section{Experiments and Results}
\label{sec:experiments}

We organize our analysis around a central question: which directions in pretrained weight matrices remain stable under finetuning, and what does this stability reveal about pretrained model transferability? We proceed in four steps: establishing that leading singular vectors resist change across diverse tasks and architectures (\textbf{Section}~\ref{subsec:layerwise_stability}); understanding why this stability arises by analyzing the spectral structure of finetuning updates(\textbf{Section}~\ref{sec:gradient}); tracing its origin to pretraining scale and data (\textbf{Section}~\ref{subsec:data_scale_results}) and exploiting it for parameter-efficient finetuning by freezing pretrained singular vectors and optimizing only spectral coefficients (\textbf{Section}~\ref{subsec:peft}).

\subsection{Pretraining Induces a Stable and Transferable Spectral Basis}
\label{subsec:layerwise_stability}

\begin{figure*}[htbp]
  \centering
  \includegraphics[width=0.98\linewidth]{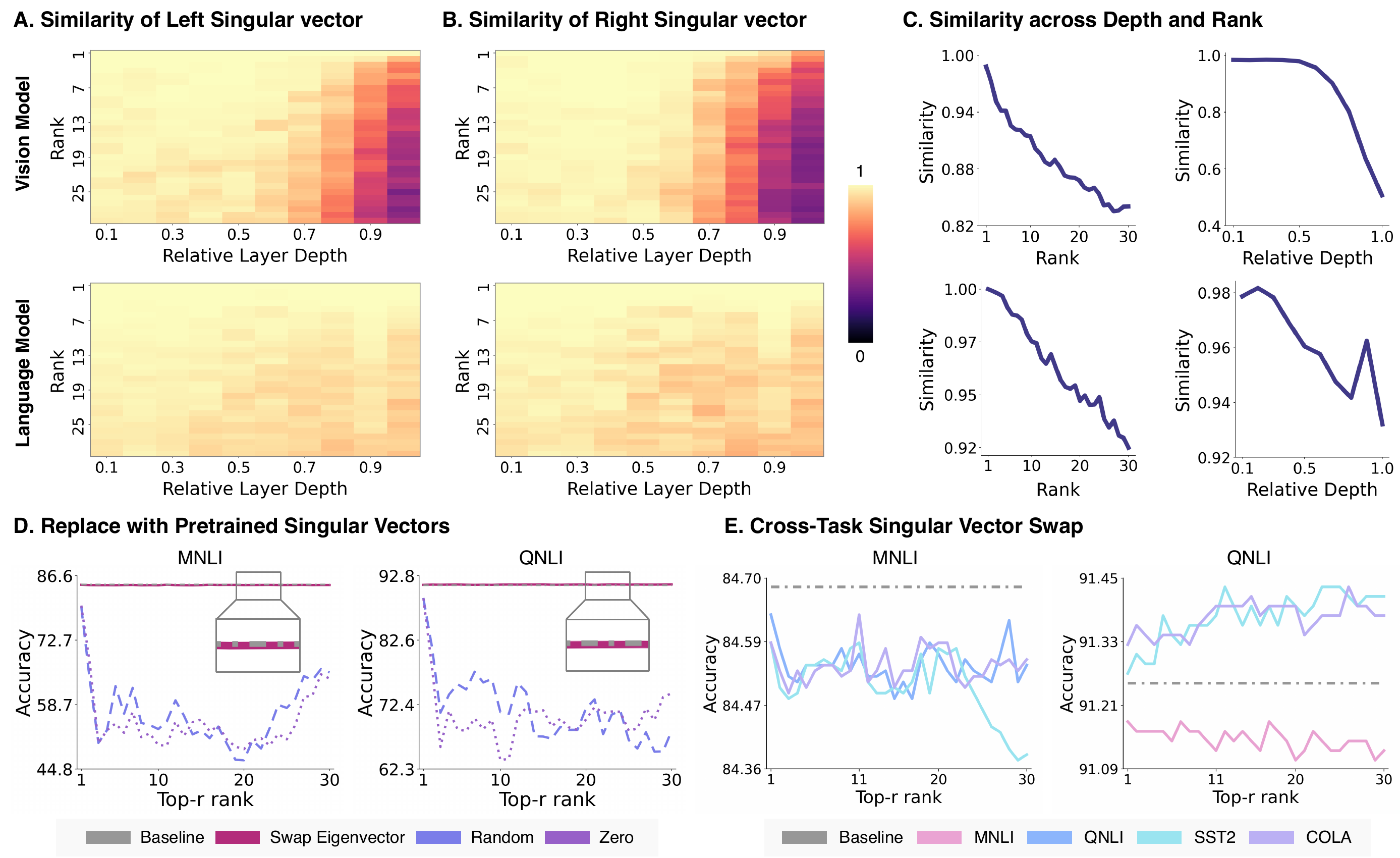}
  \caption{
  \textbf{Finetuning largely preserves the pretrained spectral basis.}
  \textbf{(A, B)} Cosine similarity between matched left/right singular vectors before and after finetuning across language and vision models.
  \textbf{(C)} Similarity as a function of rank and layer, leading directions and shallow layers are the most stable.
  \textbf{(D)} Replacing finetuned singular vectors with pretrained ones causes little performance loss.
  \textbf{(E)} Swapping singular vectors across different downstream tasks also has limited impact, while random replacement or zeroing causes clear degradation.
  }
  \label{fig:finetune}
\end{figure*}

We begin by directly comparing the singular vectors of pretrained and finetuned weight matrices across layers. Across both language and vision models, we find that the leading singular vectors remain substantially aligned after finetuning, as measured by cosine similarity between corresponding singular vectors before and after finetuning (Figure~\ref{fig:finetune}A,B).

In both modalities, alignment tends to decrease in deeper layers and for lower-ranked components (Figure~\ref{fig:finetune}C). Nevertheless, the similarity remains high overall, indicating that finetuning preserves much of the pretrained spectral structure. 

We further verify this by replacing finetuned singular vectors with pretrained ones while keeping singular values fixed. We analyze \texttt{bert-base-uncased} finetuned on MNLI and QNLI respectively, and as shown in Figure~\ref{fig:finetune}D, this substitution causes negligible performance change. By contrast, replacing the same singular vectors with random vectors or setting them to zero leads to substantial degradation, confirming that the stability reflects genuine structural reuse rather than mere insensitivity to perturbation.

We also perform the same analysis across models finetuned on different 
downstream tasks. Specifically, we analyze four downstream tasks in total: 
MNLI, QNLI, SST2, and COLA. Figure~\ref{fig:finetune}E shows the performance change when swapping singular vectors between models finetuned on MNLI/QNLI and models finetuned on other datasets. Remarkably, even when swapping the top 30 singular vectors, overall performance does not significantly degrade and sometimes even improves. This result suggests that pretraining already establishes singular-vector directions that are shared across downstream tasks, and that finetuning operates within this inherited spectral structure rather than constructing task-specific directions from scratch.

Additional model details are provided in Appendix~\ref{appendix:largemodel}. We further validate the consistency of our findings across additional modalities (Appendix~\ref{appendix:Multimodal}), downstream tasks (Appendix~\ref{appendix:Swapping}), and a more complete range of ranks (Appendix~\ref{appendix:MoreRank}), with results remaining consistent throughout. We also analyze singular value in Appendix~\ref{appendix:singular_value}, finding that singular values remain largely unchanged before and after finetuning across most models.

\subsection{Spectral Stability is Primarily Determined by the Magnitude of Update}
\label{sec:gradient}

The previous section established that leading singular vectors remain stable across finetuning. This stability suggests that the finetuning process differs from arbitrary optimization: the gradients must exhibit specific characteristics that preserve the pretrained spectral structure. To identify these characteristics, we analyze the cumulative parameter update by comparing its singular values and singular vectors to those of the pretrained weights, as defined in Section~\ref{sec:method_grad}.


\begin{figure}[htbp]
  \centering
  \includegraphics[width=0.99\linewidth]{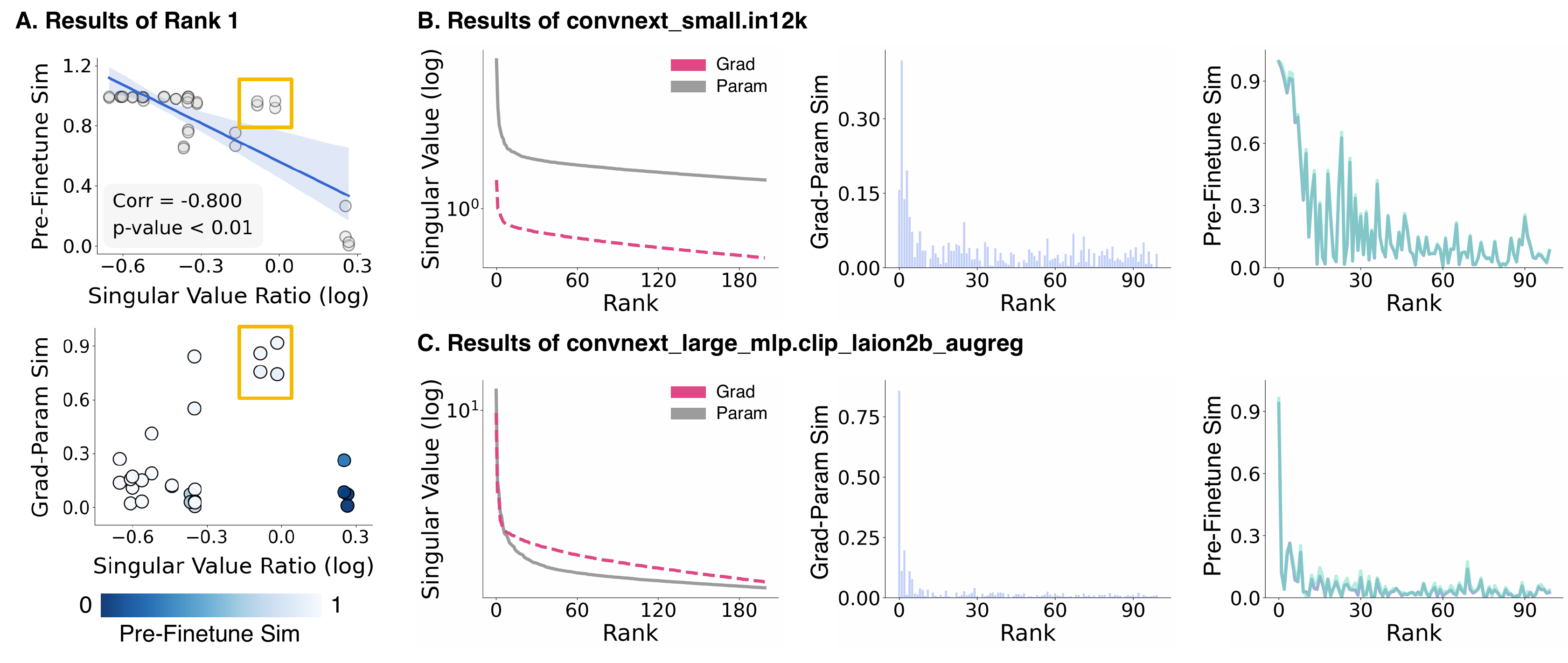}
  \caption{
  	\textbf{Spectral stability is controlled by update magnitude and directional alignment.}
  	\textbf{(A)} For the leading singular direction, smaller ratios of singular value generally correlates with higher similarity of singular vector.
  	\textbf{(B)} Representative case where $\sigma_1(\Delta W) \ll \sigma_1(W^{\text{pt}})$: the leading singular vector is preserved with high fidelity.
  	\textbf{(C)} Representative case where $\sigma_1(\Delta W) \approx \sigma_1(W^{\text{pt}})$: stability can be maintained when the update is strongly aligned with the pretrained direction.
  }
  \label{fig:gradient}
\end{figure}



\paragraph{Update magnitude is the primary driver of stability.} We quantify the magnitude of finetuning updates using the singular value ratio $r_i = \sigma_i(\Delta W) / \sigma_i(W^{\text{pt}})$, which measures the relative scale of change at each rank. Figure~\ref{fig:gradient}A shows the results of top singular vector, that smaller singular value ratios strongly correlate with higher cosine similarity between pretrained and finetuned singular vectors: when updates are small relative to pretrained weights, the spectral basis is naturally preserved.

\paragraph{Directional alignment provides a secondary explanation.} However, singular value ratios alone do not fully explain spectral stability. We observe some cases in Figure~\ref{fig:gradient}A where $\sigma_1(\Delta W) \gtrsim \sigma_1(W^{\text{pt}})$, indicating substantial parameter change, yet the pretrained singular vectors remain stable. For these cases, we find that the singular vectors of $\Delta W$ are strongly aligned with those of $W^{\text{pt}}$. This indicates that optimization proceeds along the pretrained leading singular directions, so the finetuned parameters inherit the pretrained spectral structure despite large update magnitudes.

\paragraph{Two representative cases illustrate these findings.} Figure~\ref{fig:gradient}B shows a case where $\sigma_1(\Delta W) \ll \sigma_1(W^{\text{pt}})$: the small singular value ratio directly preserves the leading singular vector. Figure~\ref{fig:gradient}C presents the complementary scenario where $\sigma_1(\Delta W) \approx \sigma_1(W^{\text{pt}})$: despite the large singular value ratio, stability is maintained because the singular vectors of $\Delta W$ are strongly aligned with the pretrained directions.

Together, these results reveal that spectral stability emerges from two complementary factors: small singular value ratios provide the primary explanation, while alignment between the singular vectors of $\Delta W$ and $W^{\text{pt}}$ explains stability even under larger updates. Additional results across more ranks and examples are provided in Appendix~\ref{appendix:Gradient}.

\subsection{Larger Pretraining Datasets Produce More Transferable Spectral Bases}
\label{subsec:data_scale_results}

The previous sections establish that finetuning reuses the pretrained 
singular-vector basis, and that stability is closely tied to the relative 
magnitude of parameter updates. It is well known that larger pretraining 
datasets lead to better model performance across a wide range of downstream tasks. But does data scale also affect the spectral structure of the pretrained weights? Specifically, we ask whether models trained on more data learn singular vector that are more transferable. To investigate this, we conduct controlled experiments varying pretraining data scale while holding the architecture and training procedure fixed, and examine how the resulting spectral basis responds to different forms of post-training adaptation.

\begin{figure*}[htbp]
  \centering
  \includegraphics[width=0.98\linewidth]{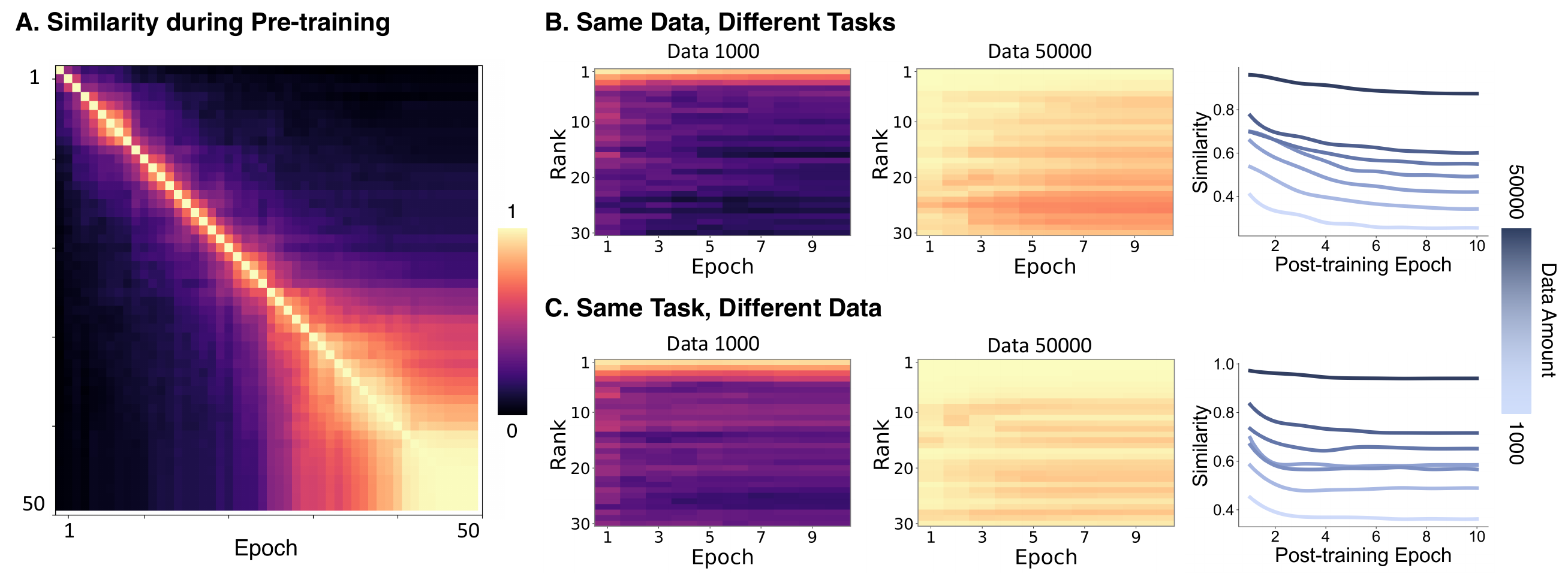}
    \caption{
    \textbf{Pretraining progressively stabilizes the spectral basis, and larger datasets make it more transferable.}
    \textbf{(A)} Cosine similarity between top singular vectors from different pretraining checkpoints. Similarity increases in later stages, showing that leading spectral directions gradually converge during pretraining.
    \textbf{(B)} Post-training on different tasks with the same data source.
    \textbf{(C)} Post-training on different data with the same task.
    In both settings, models pretrained on larger datasets exhibit smaller singular-vector changes, indicating that larger-scale pretraining produces a more stable and transferable spectral basis.
    }
  \label{fig:dataInfluence}
\end{figure*}

To test this hypothesis, we train a Vision Transformer on a masked image reconstruction task and extract the FFN parameters from the final block for analysis. We track how the leading singular vectors evolve by computing cosine similarity between checkpoints at successive epochs. As shown in Figure~\ref{fig:dataInfluence}A, leading singular vectors shift considerably in early training but progressively stabilize, with inter-checkpoint similarity increasing as training proceeds. This confirms that a stable spectral basis emerges during pretraining itself, prior to any downstream adaptation.

We then examine whether pretraining data scale affects how transferable this basis is. Using the same architecture and training procedure but varying the amount of pretraining data, we apply two forms of post-training: classification on the same data source (Figure~\ref{fig:dataInfluence}B), and masked reconstruction on a held-out dataset (Figure~\ref{fig:dataInfluence}C). In both settings, singular-vector similarity between pre- and post-training decreases as adaptation proceeds. Crucially, models pretrained on larger datasets retain higher similarity at convergence, regardless of whether the shift is in task or data distribution. This indicates that larger-scale pretraining yields a more transferable spectral basis. Additional details and results across more pretraining data scales are provided in Appendix~\ref{appendix:pretrain data}.

\subsection{Exploiting Pretrained Spectral Coordinates for Parameter-Efficient Transfer}
\label{subsec:peft}

Our spectral analysis reveals that pretrained singular vectors transfer across tasks, trailing components contribute negligibly due to small singular values, and task-specific changes concentrate in deeper layers. These findings motivate our parameter-efficient method (Section~\ref{sec: SRF}), which freezes pretrained singular vectors and optimizes only within the top-$K$ spectral subspace of selected deep layers.

\begin{figure*}[htbp]
  \centering
  \includegraphics[width=0.95\linewidth]{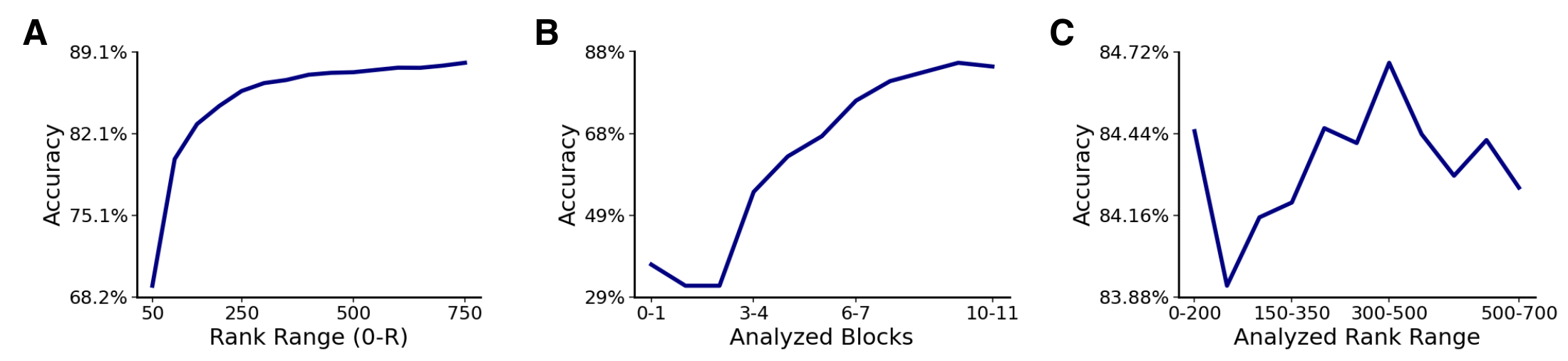}
  \caption{
  \textbf{Impact of Components on Finetuning.}
    \textbf{(A)} Increasing the range of rank consistently improves downstream performance, indicating that larger spectral subspaces capture more transferable information.
    \textbf{(B)} Optimization in deeper layers leads to better performance, highlighting the importance of deep-layer adaptation.
    \textbf{(C)} Performance differences across leading, middle, and trailing rank ranges are modest.
  }
  \label{fig:peft}
\end{figure*}

We first examine the key hyperparameters underlying this approach. First, increasing the number of optimized ranks $K$ consistently improves performance until saturation around $K=300$ (Figure~\ref{fig:peft}A), confirming that a richer spectral subspace provides more flexibility for task-specific adaptation, though further expansion yields diminishing returns. Second, when only a fixed fraction of layers can be adapted, allocating this budget to deeper layers yields substantially better results (Figure~\ref{fig:peft}B), validating that deeper parameters play a more critical role in downstream adaptation. Third, we compare different rank selection strategies for a fixed subspace size (Figure~\ref{fig:peft}C). Performance differences across leading, middle, and trailing rank ranges are modest, indicating that the choice of which specific ranks to optimize matters less than the total subspace dimensionality and layer selection.

\begin{table*}[htbp]
\centering
\small
\setlength{\tabcolsep}{5pt}
\caption{Comparison with representative PEFT methods on the GLUE benchmark using DeBERTa-v3-base. LoRA-K and LoRA-G denote LoRA with Kaiming and Gaussian initialization for B respectively.}
\label{tab:peft_baselines}
\begin{tabular}{lcccccccccc}
\toprule
Method & Params & RTE & QQP & CoLA & MRPC & SST-2 & QNLI & MNLI & STS-B & ALL \\
\midrule
Full FT   & 184M   & 83.75 & 92.40 & 69.19 & 89.46 & 95.63 & 94.03 & 89.90 & 91.60 & 88.25 \\
PiSSA \citep{meng2024pissa}     & 1.33M  & 88.69 & 92.33 & 73.12 & 91.50 & 96.22 & 94.43 & 90.37 & 92.00 & 89.83 \\
LoRA-K \citep{hu2022lora}    & 1.33M  & 84.84 & 92.03 & 70.69 & 90.28 & 95.64 & 93.84 & 89.96 & 91.68 & 88.62 \\
LoRA-G \citep{hu2022lora}    & 1.33M  & 85.20 & 91.99 & 69.82 & 89.95 & 94.95 & 93.87 & 90.65 & 91.60 & 88.50 \\
AdaLoRA \citep{zhang2023adalora}   & 1.27M  & 88.09 & 92.23 & 71.45 & 90.69 & 96.10 & 94.55 & 90.76 & 91.84 & 89.46 \\
DoRA \citep{liu2024dora}      & 1.27M  & 86.04 & 92.07 & 70.85 & 90.93 & 95.79 & 94.10 & 90.29 & 91.79 & 88.98 \\
HAdapter \citep{houlsby2019parameter} & 1.22M  & 84.48 & 91.91 & 68.64 & 89.95 & 95.53 & 94.11 & 90.13 & 91.48 & 88.28 \\
PAdapter \citep{pfeiffer2021adapterfusion}  & 1.18M  & 85.20 & 92.04 & 68.77 & 89.46 & 95.61 & 94.29 & 90.33 & 91.54 & 88.41 \\
BitFit \citep{zaken2022bitfit}    & 0.1M   & 78.70 & 88.41 & 66.96 & 87.75 & 94.84 & 92.24 & 89.37 & 91.35 & 86.20 \\
\midrule
Our method  & 0.36M & 84.12 & 89.22 & 68.13 & 88.48 & 95.53 & 92.06 & 87.45 & 88.85 & 86.73 \\
\bottomrule
\end{tabular}
\end{table*}

Based on the results above, we configure our method with 
$K=300$ applied to the final two transformer blocks and compare against representative PEFT baselines on the GLUE benchmark using DeBERTa-v3-base. Table~\ref{tab:peft_baselines} shows that our approach achieves 86.73 average score with only 0.36M trainable parameters (0.2\% of the full model), outperforming BitFit (86.20) which adapts bias terms from a different perspective. While methods like PiSSA achieve higher performance with 3--4$\times$ more parameters, our results demonstrate that a compact spectral subspace can effectively support downstream adaptation with minimal parameter overhead. This validates our central thesis: the structure of parameter established during pretraining already encodes sufficient transferable information to enable reasonable downstream adaptation. Additional experimental details are provided in Appendix~\ref{appendix:peft_details}.





\section{Discussion}

Our results reveal that pretrained models encode a reusable spectral basis that remains stable across diverse downstream tasks, reflecting a fundamental property of pretraining rather than an artifact of optimization or architecture.

\paragraph{Spectral stability reflects learned structure, not imposed constraints.}
A critical distinction underlies our findings: the observed spectral stability indicates that models \emph{do not need} to rotate their singular vectors during finetuning, rather than that they \emph{cannot}. The pretrained basis already captures the directions sufficient for downstream tasks, suggesting pretraining discovers a general-purpose coordinate system that adaptation operates within, not around. As further evidence, Appendix \ref{appendix:Failure} shows that when downstream labels are randomized, the singular vectors change substantially, confirming that this stability is not an inherent property of optimization but a consequence of meaningful task structure. Furthermore, we investigate whether the degree of this spectral stability correlates with downstream performance, but do not observe a consistent and significant relationship (Appendix \ref{appendix:generalization}). We attribute this to the substantial heterogeneity in finetuning strategies, which introduces confounding factors that obscure a clean empirical connection.


\paragraph{Generality across architectures and modalities.}
Appendix \ref{appendix:Multimodal} extends our analysis to neural signal foundation models (EEG, fMRI), with spectral stability persisting across all settings. Attention modules exhibit similar preservation as FFN layers, suggesting this is a universal property of pretrained parameters (Appendix \ref{appendix:Attention}).

\paragraph{Spectral adaptation as validation, not novelty.}
Our parameter-efficient method (Section~\ref{subsec:peft}) is conceptually similar to SVFT~\citep{lingam2024svft}, both freezing singular vectors and adapting spectral coefficients. Our contribution is empirical validation: spectral stability translates into effective transfer with 0.2\% trainable parameters, confirming pretrained singular vectors form a reusable basis.

\paragraph{Limitations.}
Our main analysis focuses on the top-30 singular vectors for clarity, with extended results up to top-200 in Appendix~\ref{appendix:MoreRank} to verify that the same trends hold over a broader rank range. To quantify how much information is captured at different ranks, we analyze explained variance in Appendix~\ref{appendix:singular_value_variance} and find that approximately 60\% of the maximum possible rank is required to explain 90\% of the spectral energy. However, as shown in Appendix~\ref{appendix:mask order}, masking leading singular vectors causes immediate performance degradation, whereas masking from the bottom can remove up to 90\% of trailing singular vectors with negligible impact. This suggests that performance-relevant information is concentrated in a small number of leading components, despite their relatively modest share of total variance. Methodologically, replacing an individual singular vector with a random vector generally destroys the orthogonality of the singular-vector basis, so the resulting perturbation is not confined to a single direction, making isolated causal attribution challenging. Finally, performing SVD on large parameter matrices incurs non-negligible computational cost, although this is a one-time initialization overhead.

\paragraph{Future directions.}
While our results show that larger pretraining corpora yield more transferable spectral bases, how to enhance spectral transferability under fixed data budgets remains an important open question. Developing pretraining objectives or architectural inductive biases that explicitly encourage the learning of more reusable singular vectors could improve model performance without requiring additional data scale.

\section*{Author contributions}
J.Y., Y.W., and Q.L. conceived and designed the study. Y.W. carried out the analyses of singular vectors and singular values for vision and language models in Section 4.1 and appendix, as well as the analysis of the impact of pre-training data scale in Section 4.3. J.Y. conducted the gradient analysis in Section 4.2. Z.D. was responsible for the analysis of PEFT in Section 4.4. Y.Z. performed the supplementary analyses of attention modules and singular vectors presented in the Appendix. W.M. handled the preprocessing of the pre-trained models. J.Y. and Y.W. drafted the manuscript with help of Q.L. All authors critically revised the manuscript and approved the final version.

\bibliography{neurips_2026}
\bibliographystyle{unsrt}


\appendix

\newpage

\section*{Appendix Overview}

This appendix provides additional empirical results, model details, and implementation specifications that further support the main claim of the paper: finetuning largely operates within a stable spectral subspace established by pretraining.

The supplementary material is organized around five broad questions.

First, \textbf{how general is spectral transfer across model families, modalities, parameter classes, and training scales?}
To address this question, we summarize the pretrained models used in our large-scale analyses and extend the singular-vector alignment study beyond standard vision and language models to EEG and fMRI models. We also examine attention modules, showing that spectral stability is not limited to FFN layers. In addition, we provide expanded results on pretraining data scale, showing that larger pretraining datasets lead to more stable singular vectors during post-training.

Second, \textbf{is spectral stability confined to only a few dominant directions, or does it persist over broader rank ranges?}
We therefore report additional rank-wise alignment analyses and variance-based results across multiple modalities and model families. These results show a consistent rank-dependent pattern: leading singular vectors are the most stable, capture a disproportionate amount of functional importance, and play a central role in downstream performance.

Third, \textbf{how do singular vector and singular value change during finetuning?}
Beyond singular-vector alignment, we include complementary analyses of singular-value spectra and masking experiments over different rank ranges. Across most pretrained models, finetuning largely preserves the singular value, with changes appearing mainly as mild rescaling. At the same time, masking analyses show that leading spectral components have a disproportionately large effect on downstream task performance.

Fourth, \textbf{when does spectral transfer break down, and what does this reveal about task compatibility?}
We study controlled failure cases, including random-label finetuning, where optimization continues from the same pretrained initialization but meaningful task structure is removed. In these settings, singular-vector alignment deteriorates substantially, indicating that spectral stability depends on compatibility between pretrained representations and downstream objectives rather than arising as a trivial consequence of continued training.

Finally, \textbf{what additional evidence shows that finetuning primarily reuses a pretrained spectral basis rather than learning entirely new directions?}
To answer this, we provide full results for singular-vector swapping, spectral similarity versus generalization, and expanded cumulative-gradient analyses. Together, these results show that finetuning updates are small relative to the pretrained spectral scale, concentrated in leading pretrained spectral coordinates, and highly aligned with pretrained singular vectors.

Taken together, the appendix strengthens the central interpretation of the paper: pretraining establishes a reusable spectral coordinate system, and downstream adaptation mostly proceeds by reusing and slightly rescaling this basis rather than replacing it with new directions.

\paragraph{Roadmap of the appendix.}
\begin{itemize}
    \item \textbf{Section~\ref{appendix:largemodel}: Details of Pretrained Models.} 
    Summary of the pretrained vision and language models used in the large-scale analyses, including model families, scales, pretraining sources, and downstream evaluation settings.

    \item \textbf{Section~\ref{appendix:singular_value}: Results of Singular Values.} 
    Complementary analysis of singular-value spectra before and after finetuning, showing that finetuning preserves the distribution of singular values and mainly introduces mild rescaling.

    \item \textbf{Section~\ref{appendix:Multimodal}: Additional Multimodal Evidence for Spectral Transfer.} 
    Extension of the singular-vector alignment analysis to EEG and fMRI models, showing that spectral transfer also appears beyond standard vision and language domains.

    \item \textbf{Section~\ref{appendix:MoreRank}: Spectral Alignment over Broader Rank Ranges.} 
    Broader rank-wise alignment analysis across vision, language, EEG, and fMRI models, demonstrating that leading spectral directions are consistently more stable than lower-ranked components.

    \item \textbf{Section~\ref{appendix:Failure}: Failure Cases of Spectral Transfer.} 
    Controlled random-label experiments showing that spectral stability breaks down when meaningful task structure is removed, highlighting the importance of compatibility between pretraining and downstream adaptation.

    \item \textbf{Section~\ref{appendix:Attention}: Singular-Vector Analysis in Attention Modules.} 
    Extension of the main FFN-layer analysis to query, key, and value projection matrices, showing that attention modules also preserve dominant pretrained singular vectors.

    \item \textbf{Section~\ref{appendix:Swapping}: Additional Results on Singular-Vector Swapping.} 
    Full task-level results for replacing or swapping singular vectors between pretrained and finetuned models and across different finetuned tasks, for both FFN and attention modules.

    \item \textbf{Section~\ref{appendix:singular_value_variance}: Variance Explained by Different Ranks.} 
    Analysis of how much information is captured by different rank ranges across various models.

    \item \textbf{Section~\ref{appendix:mask order}: Impact of Singular Vectors on Task Performance.} 
    Analysis of how masking singular vectors from top ranks versus bottom ranks affects model performance, demonstrating that leading components disproportionately determine downstream generalization despite representing a minority of spectral energy.

    \item \textbf{Section~\ref{appendix:generalization}: Spectral Similarity and Downstream Generalization.} Analysis of the relationship between singular-vector similarity (across layers and ranks) before and after finetuning and the model's downstream task generalization performance.

    \item \textbf{Section~\ref{appendix:Gradient}: Additional Details for the Cumulative-Gradient Analysis.} 
    Expanded cumulative-gradient results across more pretrained models and finer-grained top-$K$ analyses, showing that finetuning updates are small in magnitude and concentrated in the pretrained spectral basis.

    \item \textbf{Section~\ref{appendix:pretrain data}: Additional Results on the Effect of Pretraining Data Scale on Spectral Stability.}
    Expanded results across more pretraining data scales, showing that larger pretraining datasets lead to more stable singular vectors and a more transferable pretrained spectral basis during post-training.
    
    \item \textbf{Section~\ref{appendix:peft_details}: Implementation Details for PEFT Methods.} 
    Detailed implementation specifications for parameter-efficient finetuning algorithms used in the experiments, including hyperparameters and training configurations.

\end{itemize}

\section{Details of Pretrained Models}
\label{appendix:largemodel}

This appendix summarizes the pretrained models used in our analyses. 
The main text focuses on vision and language models, demonstrating spectral transfer across diverse architectures, parameter scales, and pretraining regimes. 
In this appendix, we provide complete model specifications for both the vision and language models analyzed in the main text, and additionally present details of the EEG and fMRI models used in our multimodal extension.

\subsection{Vision and Language Models}

For vision, we study ConvNeXt models spanning multiple model sizes and pretraining sources, including ImageNet-12K, ImageNet-22K, and CLIP-style LAION pretraining. For language, we study a diverse collection of pretrained encoders and encoder-decoder variants, ranging from compact BERT-style models to large DeBERTa models. Together, these models allow us to test whether the spectral stability observed in the main text persists across substantial variation in architecture, scale, and pretraining data.

Table~\ref{tab:imagenet_models} lists the vision models and their Top-1 accuracy on ImageNet-1K. These models cover both standard supervised pretraining and large-scale multimodal pretraining, and are used in our cross-scale spectral analyses in vision.

Table~\ref{tab:mnli_models} lists the language models and their downstream accuracy on MNLI. The collection includes BERT, ALBERT, DistilBERT, DistilBART, SciBERT, and DeBERTa variants, providing a broad range of language-model scales and pretraining recipes for our transfer analysis.

\begin{table}[htbp]
\centering
\caption{Top-1 accuracy of large-scale pretrained vision models on ImageNet-1K.}
\label{tab:imagenet_models}
\resizebox{\textwidth}{!}{%
\begin{tabular}{l l c}
\toprule
\textbf{Pretrained Model} & \textbf{Finetuned Model Name} & \textbf{Acc (\%)} \\
\midrule
convnext\_nano.in12k & convnext\_nano.in12k\_ft\_in1k & 87.86 \\
convnext\_tiny.in12k & convnext\_tiny.in12k\_ft\_in1k & 88.91 \\
convnext\_tiny.fb\_in22k & convnext\_tiny.fb\_in22k\_ft\_in1k & 88.02 \\
convnext\_small.fb\_in22k & convnext\_small.fb\_in22k\_ft\_in1k & 89.93 \\
convnext\_small.in12k & convnext\_small.in12k\_ft\_in1k & 89.03 \\
convnext\_base.clip\_laion2b\_augreg & convnext\_base.clip\_laion2b\_augreg\_ft\_in1k & 89.68 \\
convnext\_large\_mlp.clip\_laion2b\_augreg & convnext\_large\_mlp.clip\_laion2b\_augreg\_ft\_in1k & 90.35 \\
convnext\_large.fb\_in22k & convnext\_large.fb\_in22k\_ft\_in1k & 90.31 \\
convnext\_xlarge.fb\_in22k & convnext\_xlarge.fb\_in22k\_ft\_in1k & 90.10 \\
convnext\_xxlarge.clip\_laion2b\_soup & convnext\_xxlarge.clip\_laion2b\_soup\_ft\_in1k & 90.67 \\
timm/vit\_base\_patch16\_224.augreg\_in21k & timm/vit\_base\_patch16\_224.augreg\_in21k\_ft\_in1k & 84.50 \\
timm/vit\_large\_patch16\_224.augreg\_in21k & timm/vit\_large\_patch16\_224.augreg\_in21k\_ft\_in1k & 85.90 \\
timm/swin\_base\_patch4\_window7\_224.ms\_in22k & timm/swin\_base\_patch4\_window7\_224.ms\_in22k\_ft\_in1k & 85.20 \\
timm/swin\_large\_patch4\_window7\_224.ms\_in22k & timm/swin\_large\_patch4\_window7\_224.ms\_in22k\_ft\_in1k & 86.30 \\
\bottomrule
\end{tabular}%
}
\end{table}

\begin{table}[htbp]
\centering
\caption{Accuracy of pretrained language models fine-tuned on the MNLI benchmark.}
\label{tab:mnli_models}
\resizebox{\textwidth}{!}{%
\begin{tabular}{l l c}
\toprule
\textbf{Pretrained Model} & \textbf{Finetuned Model Name} & \textbf{Acc (\%)} \\
\midrule
google-bert/bert-base-uncased & ishan/bert-base-uncased-mnli & 84.60 \\
google-bert/bert-large-uncased & madlag/bert-large-uncased-mnli & 86.05 \\
google-bert/bert-base-cased & gchhablani/bert-base-cased-finetuned-mnli & 84.60 \\
google-bert/bert-large-cased & George-Ogden/bert-large-cased-finetuned-mnli & 86.09 \\
FacebookAI/roberta-base & textattack/roberta-base-MNLI & 87.60 \\
FacebookAI/roberta-large & FacebookAI/roberta-large-mnli & 90.20 \\
xlnet-base-cased & vish88/xlnet-base-mnli-finetuned & 86.80 \\
xlnet-large-cased & ynie/xlnet-large-cased-snli\_mnli\_fever\_anli\_R1\_R2\_R3-nli & 90.80 \\
microsoft/deberta-large & microsoft/deberta-large-mnli & 91.30\\
microsoft/deberta-xlarge & microsoft/deberta-xlarge-mnli & 91.50\\
microsoft/deberta-v2-xlarge & microsoft/deberta-xlarge-v2-mnli & 91.70 \\
microsoft/deberta-v2-xxlarge & microsoft/deberta-xxlarge-v2-mnli & 91.90 \\
microsoft/deberta-v3-base & MoritzLaurer/DeBERTa-v3-base-mnli & 90.60 \\
microsoft/deberta-v3-large & khalidalt/DeBERTa-v3-large-mnli & 91.80 \\
distilbert/distilbert-base-uncased & typeform/distilbert-base-uncased-mnli & 88.80 \\
google/bigbird-roberta-base & l-yohai/bigbird-roberta-base-mnli & 87.50 \\
google/mobilebert-uncased & Alireza1044/mobilebert\_mnli & 84.00\\
google/electra-base-discriminator & howey/electra-base-mnli & 82.20 \\
\bottomrule
\end{tabular}%
}
\end{table}

\subsection{EEG and fMRI Models}

We analyze one EEG foundation model and two fMRI foundation models to validate the generality of our findings across brain signal modalities.

\paragraph{EEG Model: LaBraM.}
We adopt LaBraM-Base~\citep{jiang2024large} as the EEG foundation model.
LaBraM first employs vector-quantized neural spectrum prediction to train a
semantically rich neural tokenizer, which encodes continuous raw EEG channel
patches (200 sampling points per patch at 200\,Hz, corresponding to 1-second
segments) into compact discrete neural codes.
Neural Transformers are then pre-trained via masked prediction, where the model
predicts the original neural codes of masked EEG channel patches.
Pre-training is conducted on approximately 2{,}500 hours of EEG data spanning
around 20 datasets, using the AdamW optimizer ($\beta_1{=}0.9$, $\beta_2{=}0.98$)
with a learning rate of $5{\times}10^{-4}$, gradient clipping threshold of 3.0,
for 50 epochs.
For fine-tuning, we adopt full fine-tuning on the COG-BCI
dataset~\citep{hinss2023open}, combining MATB and N-Back tasks for a six-class
classification.
Data is split into train/validation/test sets at a 70\%/10\%/20\% ratio.
Training uses a learning rate of $1{\times}10^{-4}$, batch size of 8, and 300
epochs, with random seed fixed at 42.
The model employs absolute positional embeddings, with QKV bias terms and
relative positional bias disabled.

\paragraph{fMRI Model: BrainLM.}
BrainLM-Base~\citep{carobrainlm} (111M parameters) is an fMRI foundation
model based on the Masked Autoencoder (MAE) architecture, pre-trained with
self-supervised masked prediction on 6{,}700 hours of resting-state fMRI
recordings from the UK Biobank (UKB) and the Human Connectome Project (HCP).
For fine-tuning~\citep{wang2026omni}, we use the AdamW optimizer
($\beta_1{=}0.9$, $\beta_2{=}0.999$, $\varepsilon{=}1{\times}10^{-8}$,
weight decay$\,{=}\,0.01$) with a learning rate of $5{\times}10^{-4}$,
batch size of 16, and 100 training epochs.
Each fMRI sample is uniformly truncated to 200 time steps as model input,
and the random seed is fixed at 51.

\paragraph{fMRI Model: Brain-JEPA.}
Brain-JEPA-Base~\citep{dong2024brain} (ViT-Base, 86M parameters) adopts the
Joint-Embedding Predictive Architecture (JEPA) for brain dynamics modeling.
The model is pre-trained for 300 epochs on resting-state fMRI data from the
UK Biobank dataset, comprising 40{,}162 participants aged 44--83 years.
For fine-tuning~\citep{xia2026brain}, we load the checkpoint from epoch 300
and adapt the model to an age regression task on the NKI dataset.
The input consists of 450 ROIs $\times$ 160 time steps (downsampling rate of 3),
with patch size 16.
We use a base learning rate of $1{\times}10^{-3}$ with cosine decay to a minimum
learning rate of $1{\times}10^{-6}$, batch size of 8, and 50 training epochs.
Regression targets are normalized using training-set Z-score normalization, and
a trainable linear mapping head is appended for regression prediction.
Attention is computed using Flash Attention for efficiency.
All results are averaged over 5 different random seeds to ensure reliability.

\section{Results of Singular Values}
\label{appendix:singular_value}

The main text focuses on the stability of singular vectors during finetuning. Here we provide a complementary analysis of singular values. For each pretrained weight matrix and its finetuned counterpart, we compute the singular values and compare their changes after finetuning. Overall, we find that the singular-value spectra are also highly stable for most models: finetuning typically introduces only mild rescaling of the pretrained spectrum, rather than substantially reshaping it.

\begin{figure*}[htbp]
  \centering
  \includegraphics[width=0.98\linewidth]{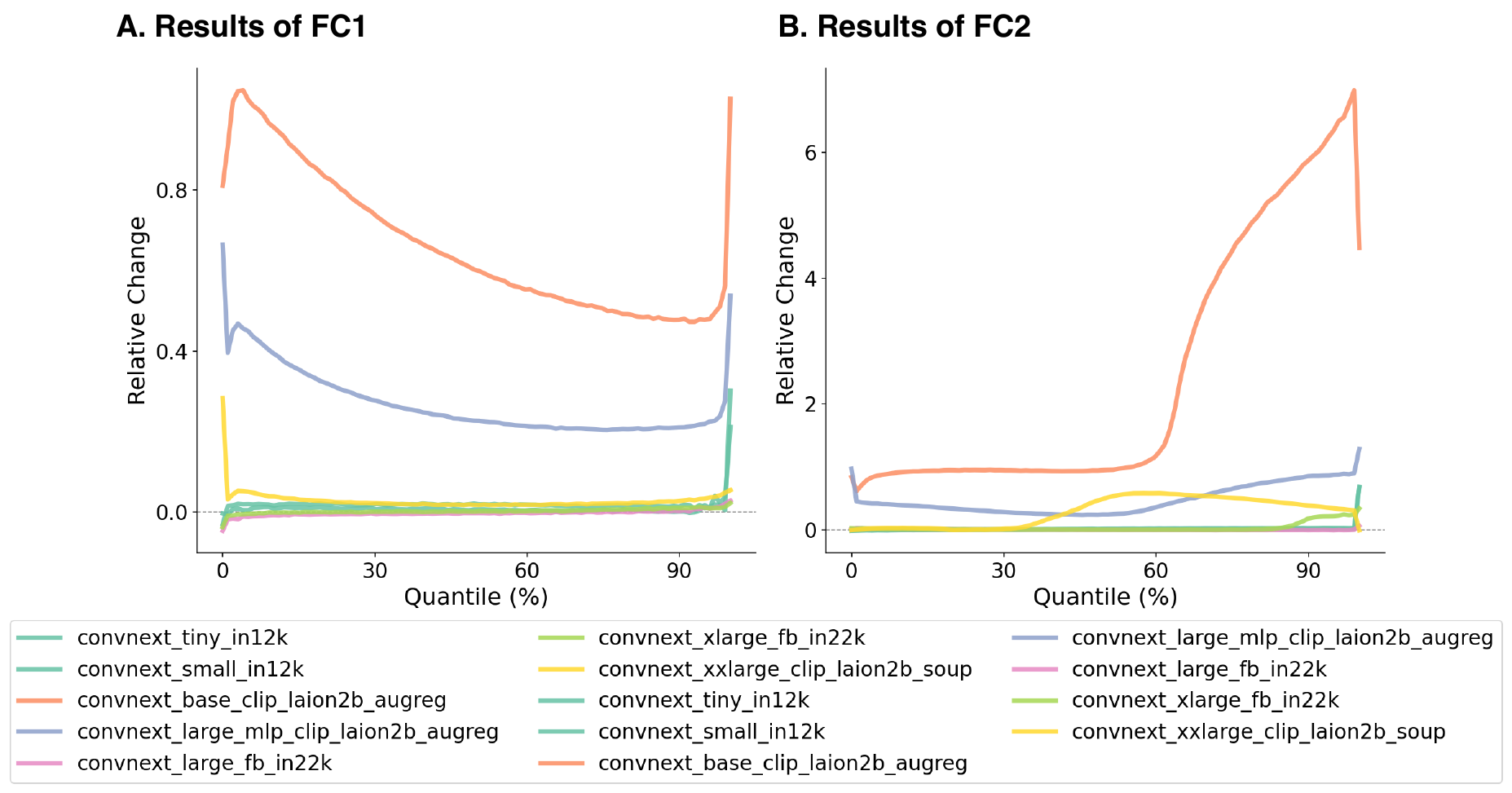}
  \caption{
  \textbf{Singular-value spectra before and after finetuning.}
  The pretrained and finetuned spectra are highly similar for most models, while larger spectral rescaling mainly appears in models with stronger pretraining--finetuning distribution shifts, such as multimodal pretrained models adapted to ImageNet.
  }
  \label{fig:Appendix_SingularValue}
\end{figure*}

Figure~\ref{fig:Appendix_SingularValue} shows the relative change in singular values across different ranks for models in our analysis. For each model and rank, we compute the relative change as the difference between finetuned and pretrained singular values divided by the pretrained value, visualized along the quantile axis. In most cases, the relative changes remain small across all ranks, indicating that finetuning preserves both the leading singular vectors and their spectral magnitudes. Noticeable deviations appear only in a small subset of models. Notably, these exceptions are primarily vision-language multimodal models, where adaptation may require stronger rescaling of pretrained features.

To quantify these observations, Table~\ref{tab:sv_analysis} reports spectral change statistics for the last-block MLP weights of models at different scales. We use the relative spectral distance,
$\mathrm{RSD}
=
\frac{\|\mathbf{s}_{\mathrm{ft}}-\mathbf{s}_{\mathrm{pre}}\|_2}
{\|\mathbf{s}_{\mathrm{pre}}\|_2},$
to measure the overall change in the singular-value vector, where $\mathbf{s}_{\mathrm{pre}}$ and $\mathbf{s}_{\mathrm{ft}}$ denote the pretrained and finetuned singular-value vectors, respectively. 

We further report the mean relative change (MRC) and maximum relative change (MaxRC) to summarize elementwise variation, as well as the top-1 relative change (T1RC) and tail relative change (TailRC) to separate the behavior of the largest singular value from that of the low-magnitude tail. TailRC is computed over the last $10\%$ singular values. 

Finally, we report the effective rank (ER) before and after finetuning, which measures the spread of the normalized singular-value distribution. A positive $\Delta$ER indicates a more spread-out spectrum after finetuning.

The results show that most models exhibit very small spectral changes, the RSD values are generally below $0.03$, and the changes in effective rank are also minor. These results indicate that the singular-value distribution is nearly preserved after finetuning. 

In contrast, a few models display much larger spectral changes. ConvNeXt-Base and the ConvNeXt-Large variant have substantially higher RSD, MRC, and MaxRC values, indicating stronger anisotropic rescaling of the spectrum. Notably, these models with significant spectral changes are all vision-language multimodal models. 

Taken together, the singular-value analysis supports the conclusion that finetuning is spectrally constrained. For most models, finetuning leaves the pretrained singular-value spectrum almost unchanged.

\begin{table}[t]
\centering
\caption{
\textbf{Spectral change statistics for last-block MLP weights across ConvNeXt, ViT, and Swin models.}
Most models show small relative spectral distance after finetuning, while CLIP-pretrained ConvNeXt models exhibit substantially stronger singular-value rescaling. ViT and Swin models show minimal spectral change.
}
\label{tab:sv_analysis}
\resizebox{\linewidth}{!}{
\begin{tabular}{lll r rrrrr rrr}
\toprule
\multirow{2}{*}{Arch} & \multirow{2}{*}{Model} & \multirow{2}{*}{Layer} & \multirow{2}{*}{$n_{\text{sv}}$}
  & \multicolumn{5}{c}{Singular Value Change Metrics}
  & \multicolumn{3}{c}{Effective Rank} \\
\cmidrule(lr){5-9} \cmidrule(lr){10-12}
 & & & & RSD & MRC & MaxRC & T1RC & TailRC
 & ER$_{\text{pre}}$ & ER$_{\text{ft}}$ & $\Delta$ER \\
\midrule
\multirow{16}{*}{\rotatebox[origin=c]{90}{ConvNeXt}}
& \multirow{2}{*}{CNX-Tiny (in12k)}
  & fc1 &  768 & 0.0089 & 0.0089 & 0.2108 & $-$0.0033 & $+$0.0158 &  725.7 &  726.1 &  $+$0.3 \\
& & fc2 &  768 & 0.0077 & 0.0147 & 0.6706 & $-$0.0131 & $+$0.0150 &  743.9 &  745.2 &  $+$1.3 \\
& \multirow{2}{*}{CNX-Tiny (in22k)}
  & fc1 &  768 & 0.0119 & 0.0137 & 0.1526 & $-$0.0215 & $-$0.0224 &  735.6 &  733.4 &  $-$2.2 \\
& & fc2 &  768 & 0.0314 & 0.0324 & 0.8587 & $+$0.0226 & $+$0.0151 &  737.3 &  735.6 &  $-$1.7 \\
& \multirow{2}{*}{CNX-Small (in22k)}
  & fc1 &  768 & 0.0107 & 0.0078 & 0.0797 & $+$0.0043 & $+$0.0005 &  726.9 &  727.5 &  $+$0.7 \\
& & fc2 &  768 & 0.0188 & 0.0226 & 0.8908 & $+$0.1028 & $-$0.0028 &  739.1 &  737.7 &  $-$1.4 \\
& \multirow{2}{*}{CNX-Small (in12k)}
  & fc1 &  768 & 0.0186 & 0.0201 & 0.3012 & $-$0.0307 & $+$0.0161 &  730.9 &  731.2 &  $+$0.3 \\
& & fc2 &  768 & 0.0125 & 0.0208 & 0.6847 & $-$0.0138 & $+$0.0285 &  736.6 &  738.8 &  $+$2.3 \\
& \multirow{2}{*}{CNX-Base (CLIP)}
  & fc1 & 1024 & 0.7734 & 0.6639 & 1.0480 & $+$0.8108 & $+$0.4978 &  954.5 &  922.8 & $-$31.6 \\
& & fc2 & 1024 & 1.0209 & 2.4158 & 6.9866 & $+$0.8395 & $+$6.3601 &  779.2 &  925.4 & $+$146.2 \\
& \multirow{2}{*}{CNX-Large-MLP (CLIP)}
  & fc1 & 1536 & 0.3304 & 0.2709 & 0.6640 & $+$0.6640 & $+$0.2310 & 1445.6 & 1417.6 & $-$28.0 \\
& & fc2 & 1536 & 0.3789 & 0.4771 & 1.2856 & $+$0.9725 & $+$0.8735 & 1183.6 & 1208.4 & $+$24.8 \\
& \multirow{2}{*}{CNX-Large (CLIP)}
  & fc1 & 1536 & 0.0110 & 0.0055 & 0.0467 & $-$0.0467 & $+$0.0093 & 1417.3 & 1420.6 &  $+$3.3 \\
& & fc2 & 1536 & 0.0100 & 0.0057 & 0.0633 & $+$0.0092 & $-$0.0012 & 1471.4 & 1469.2 &  $-$2.2 \\
& \multirow{2}{*}{CNX-XLarge (in22k)}
  & fc1 & 2048 & 0.0071 & 0.0042 & 0.0376 & $-$0.0376 & $+$0.0107 & 1855.3 & 1858.4 &  $+$3.0 \\
& & fc2 & 2048 & 0.0156 & 0.0349 & 0.3397 & $+$0.0234 & $+$0.2184 & 1776.2 & 1787.0 & $+$10.8 \\
& \multirow{2}{*}{CNX-XXL (CLIP)}
  & fc1 & 3072 & 0.0505 & 0.0280 & 0.2824 & $+$0.2824 & $+$0.0346 & 2821.1 & 2812.2 &  $-$9.0 \\
& & fc2 & 3072 & 0.0613 & 0.2843 & 0.5815 & $-$0.0051 & $+$0.3596 & 1673.7 & 1858.2 & $+$184.6 \\
\midrule
\multirow{4}{*}{\rotatebox[origin=c]{90}{ViT}}
& \multirow{2}{*}{ViT-B/16}
  & fc1 &  768 & 0.0018 & 0.0013 & 0.0220 & $+$0.0123 & $-$0.0001 &  728.8 &  728.7 &  $-$0.1 \\
& & fc2 &  768 & 0.0008 & 0.0003 & 0.0035 & $-$0.0035 & $-$0.0002 &  736.2 &  736.2 &  $-$0.0 \\
& \multirow{2}{*}{ViT-L/16}
  & fc1 & 1024 & 0.0013 & 0.0005 & 0.0117 & $+$0.0117 & $+$0.0004 &  954.9 &  954.9 &  $+$0.0 \\
& & fc2 & 1024 & 0.0007 & 0.0001 & 0.0019 & $-$0.0019 & $+$0.0002 &  966.4 &  966.4 &  $+$0.0 \\
\midrule
\multirow{4}{*}{\rotatebox[origin=c]{90}{Swin}}
& \multirow{2}{*}{Swin-B}
  & fc1 & 1024 & 0.0083 & 0.0086 & 0.0156 & $+$0.0035 & $+$0.0091 &  945.3 &  945.5 &  $+$0.2 \\
& & fc2 & 1024 & 0.0074 & 0.0080 & 0.0364 & $+$0.0084 & $+$0.0096 &  969.3 &  969.5 &  $+$0.2 \\
& \multirow{2}{*}{Swin-L}
  & fc1 & 1536 & 0.0047 & 0.0043 & 0.0089 & $+$0.0055 & $+$0.0041 & 1402.6 & 1402.4 &  $-$0.2 \\
& & fc2 & 1536 & 0.0058 & 0.0051 & 0.0198 & $+$0.0122 & $+$0.0061 & 1447.7 & 1447.8 &  $+$0.0 \\
\bottomrule
\end{tabular}
}
\end{table}

\section{Additional Multimodal Evidence for Spectral Stability}
\label{appendix:Multimodal}

The main text studies singular-vector alignment primarily in vision and language models. 
To examine whether the same phenomenon extends beyond these two domains, we repeat the analysis on pretrained models from two additional modalities: EEG and fMRI.

\begin{figure*}[htbp]
  \centering
  \includegraphics[width=0.98\linewidth]{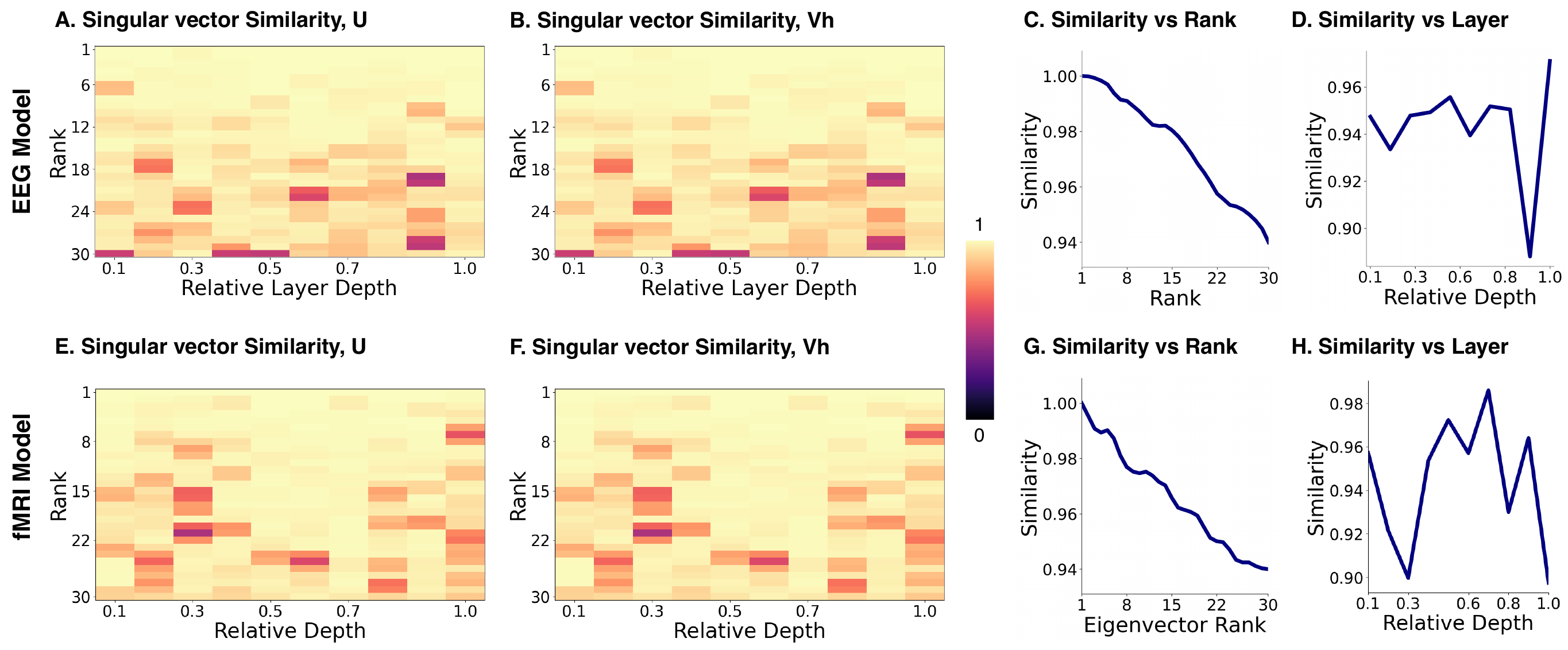}
  \caption{
  \textbf{Spectral transfer in EEG and fMRI models.}
  We repeat the singular-vector alignment analysis for pretrained EEG and fMRI models. 
  Panels \textbf{(A--D)} show EEG results, and panels \textbf{(E--H)} show fMRI results. 
  In both modalities, pretrained and finetuned singular vectors remain highly aligned, especially in the leading spectral components.
  }
  \label{fig:Appendix_multiModal}
\end{figure*}

Figure~\ref{fig:Appendix_multiModal} shows that the same qualitative pattern observed in the main text also holds for EEG and fMRI models. 
Across layers, the leading left and right singular vectors of the finetuned weights remain strongly aligned with those of the pretrained weights. 
The alignment is consistently strongest for the top-ranked spectral components and gradually weakens for lower-ranked directions, indicating that finetuning preserves the dominant pretrained spectral coordinates while allowing more flexibility in less dominant components.

These results suggest that the spectral structure induced by pretraining is not specific to a particular architecture or data modality. 
Instead, pretraining appears to establish a transferable layerwise coordinate system that continues to organize downstream adaptation across substantially different domains.

\section{Spectral Stability over Broader Rank Ranges}
\label{appendix:MoreRank}

The main text focuses on the most dominant spectral directions and shows that finetuning largely preserves the pretrained spectral basis, especially in the top-ranked components. Here we extend this analysis to a broader range of singular-value ranks, averaging results across both FC1 and FC2 layers to compare models from all four modalities in a unified view.

\begin{figure*}[!t]
  \centering
  \includegraphics[width=0.98\linewidth]{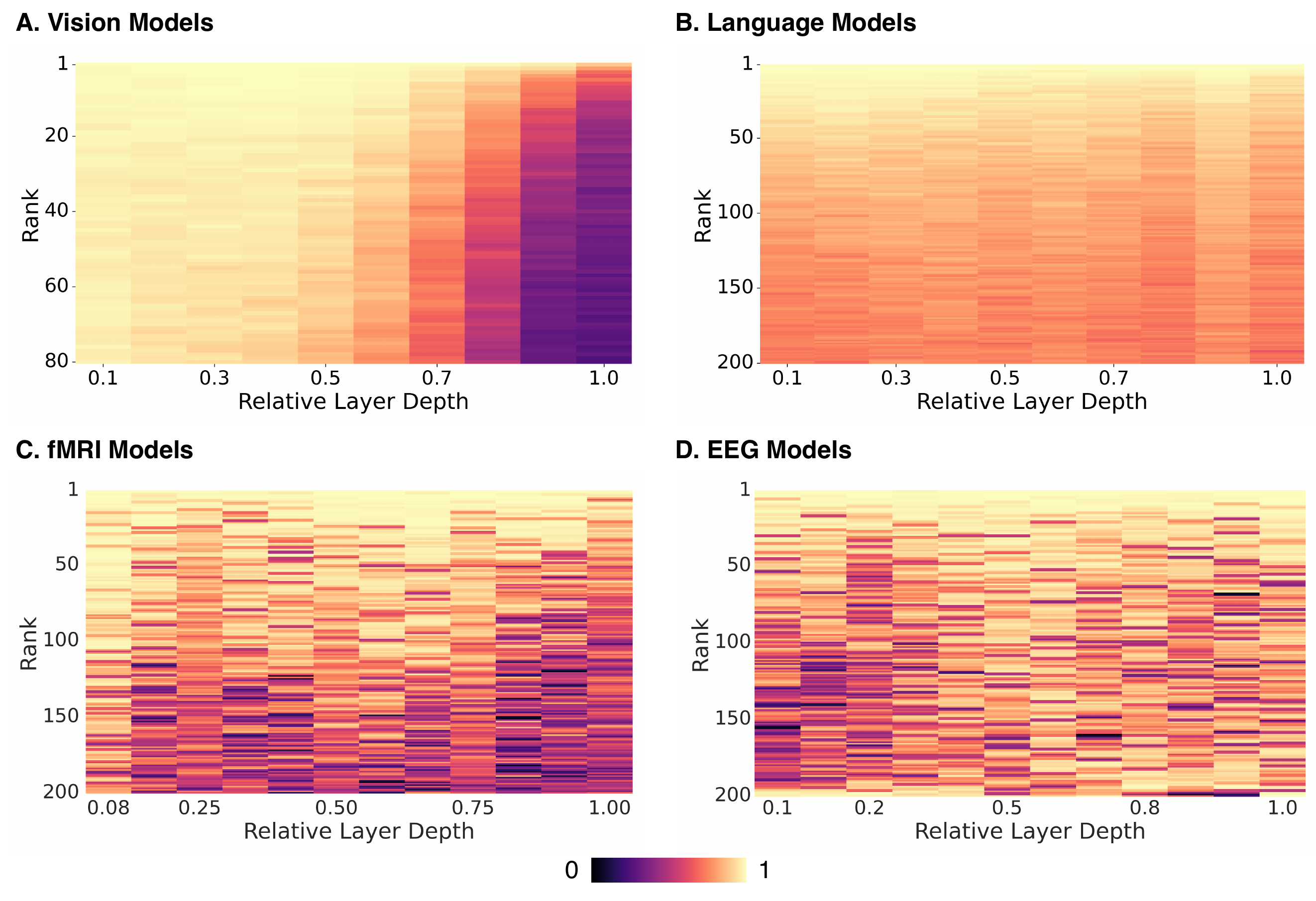}
  \caption{
  \textbf{Singular-vector alignment over broader rank ranges.}
  We compare rank-dependent alignment across vision, language, EEG, and fMRI models. 
  }
  \label{fig:Appendix_moreRank}
\end{figure*}

Figure~\ref{fig:Appendix_moreRank} shows a consistent rank-dependent pattern across vision, language, EEG, and fMRI models. Leading singular vectors remain highly similar after finetuning, while lower-ranked directions exhibit progressively larger changes. Notably, fMRI and EEG models show a more pronounced similarity drop even in shallow layers compared to vision and language models. We attribute this to the current generation of EEG and fMRI pretrained models not yet learning sufficiently transferable spectral structures, resulting in larger overall changes during finetuning and consequently less satisfactory generalization performance.

\section{Failure Cases of Spectral Stability}
\label{appendix:Failure}

The main text shows that leading singular vectors of pretrained weights remain highly stable during downstream finetuning across diverse modalities and architectures. 
To demonstrate that this phenomenon is nontrivial, we construct a setting where transfer compatibility is intentionally destroyed. 
This experiment reveals that spectral stability is not an automatic consequence of continued training, especially given the high dimensionality of modern neural networks where random alignment would be exceedingly unlikely.

We isolate the role of task structure by keeping the input data fixed while randomizing the labels. 
Using models pretrained as described in Section~\ref{subsec:data_scale_results}, we finetune them on CIFAR10 with either true labels or randomly shuffled labels. 
This preserves the data distribution while removing any meaningful semantic relation between pretraining and the downstream objective.

\begin{figure*}[!t]
  \centering
  \includegraphics[width=0.9\linewidth]{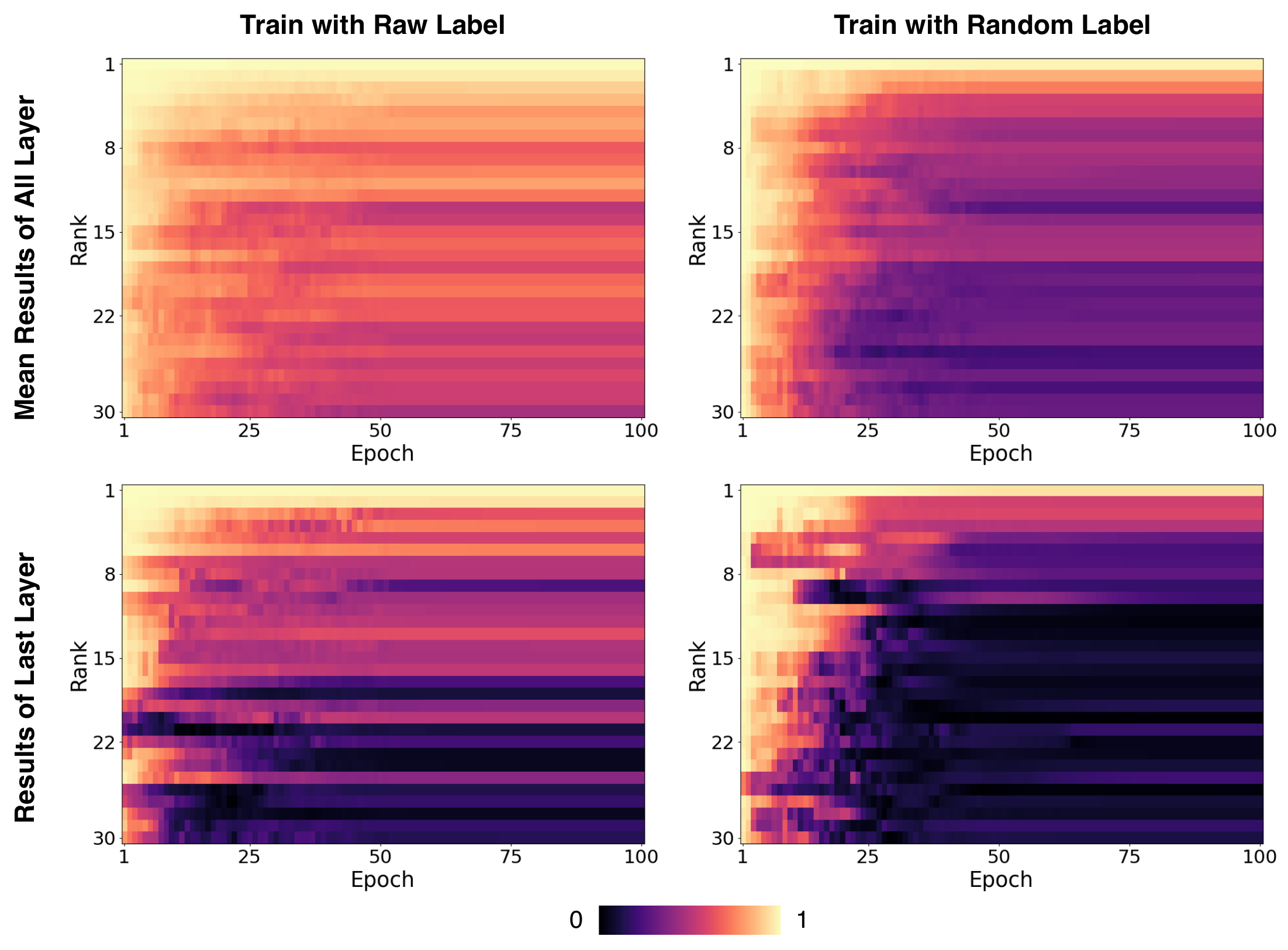}
  \caption{
    \textbf{Random labels break spectral stability.}
    Starting from the same pretrained checkpoint and using the same input data, singular vectors remain stable under true labels but rapidly lose alignment under random labels. 
  }
  \label{fig:Appendix_failureCase}
\end{figure*}

Figure~\ref{fig:Appendix_failureCase} shows that under random labels, singular vectors change much more rapidly during training. 
Cosine similarity deteriorates significantly faster compared to training with true labels, and the final similarity is substantially lower. 
This confirms that spectral stability depends critically on meaningful task alignment rather than merely sharing the same data distribution or parameter initialization.

This failure case provides strong evidence that the spectral stability observed in the main text reflects genuine transfer of learned structure, not a trivial consequence of high-dimensional geometry or continued optimization.

\section{Singular-Vector Analysis in Attention Modules}
\label{appendix:Attention}

The main text focuses on feed-forward (FFN) layers. To assess whether the same phenomenon extends to other components, we perform the same singular-vector alignment analysis on attention modules, specifically the query, key, value and output projection matrices.

\begin{figure*}[htbp]
  \centering
  \includegraphics[width=0.98\linewidth]{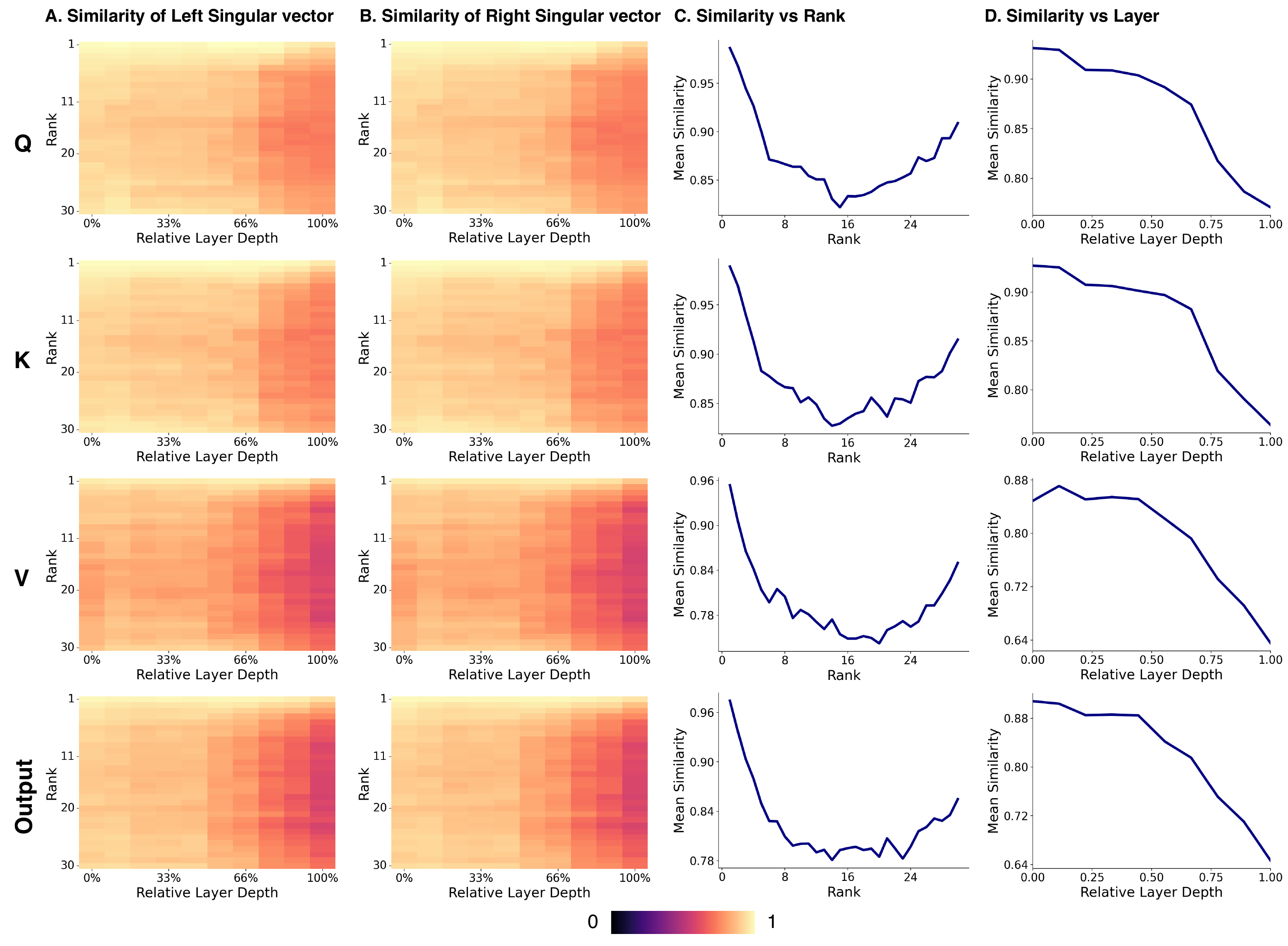}
  \caption{
    \textbf{Attention-module spectral alignment in vision models.}
    Singular-vector similarity for query, key, value and output projections shows the same  pattern as FFN layers: leading components remain highly stable after finetuning.
  }
  \label{fig:Appendix_attention_vision}
\end{figure*}

\begin{figure*}[!t]
  \centering
  \includegraphics[width=0.98\linewidth]{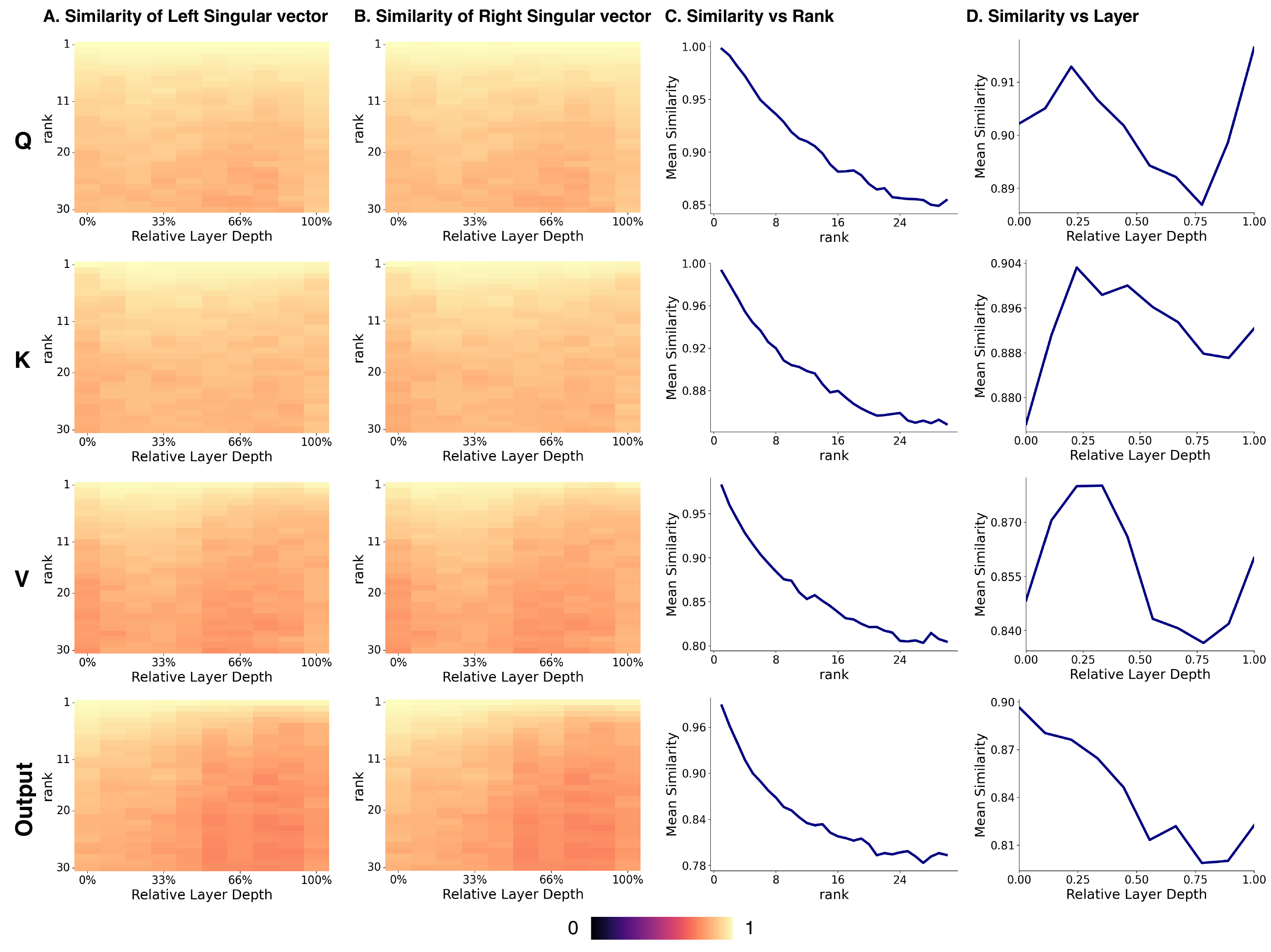}
  \caption{
    \textbf{Attention-module spectral alignment in language models.}
    Across query, key, and value projections, leading singular vectors remain strongly 
    aligned after finetuning, with alignment decreasing toward lower-ranked components, 
    consistent with FFN and vision-model observations.
  }
  \label{fig:Appendix_attention_language}
\end{figure*}

Figures~\ref{fig:Appendix_attention_vision} and~\ref{fig:Appendix_attention_language} show that attention modules exhibit a highly consistent pattern with FFN layers. Across query, key, value and output projections, the leading singular vectors remain strongly aligned after finetuning. 

A notable observation is that attention projections exhibit even higher 
alignment than FFN layers, with this pattern being particularly pronounced in vision models. This suggests that the spectral basis in attention modules is particularly stable under downstream adaptation, possibly reflecting stronger structural constraints in these components.

Taken together with the FFN results in the main text, these findings show that spectral stability is not limited to a specific layer type. Instead, pretraining induces a layerwise spectral geometry that is shared across major components of the model and continues to organize finetuning throughout the network.

\section{Additional Results on Singular-Vector Swapping}
\label{appendix:Swapping}

The main text presents representative swapping results for \texttt{bert-base-uncased} on two downstream tasks. 
Here we report the complete results across all four tasks (CoLA, MNLI, QNLI, SST-2), and extend the analysis to both feed-forward and attention modules.

\paragraph{Experimental design.}
We consider two complementary swapping settings:
(i) \emph{Pretrain $\rightarrow$ Finetune}: replacing the singular vectors of a finetuned model with those from its pretrained counterpart, and 
(ii) \emph{Finetune $\leftrightarrow$ Finetune}: swapping singular vectors between models finetuned on different tasks. In both cases, singular values are kept fixed. The first setting tests whether finetuning significantly alters the pretrained spectral basis, while the second tests whether different tasks require task-specific singular vectors.

\begin{figure*}[!t]
  \centering
  \includegraphics[width=0.98\linewidth]{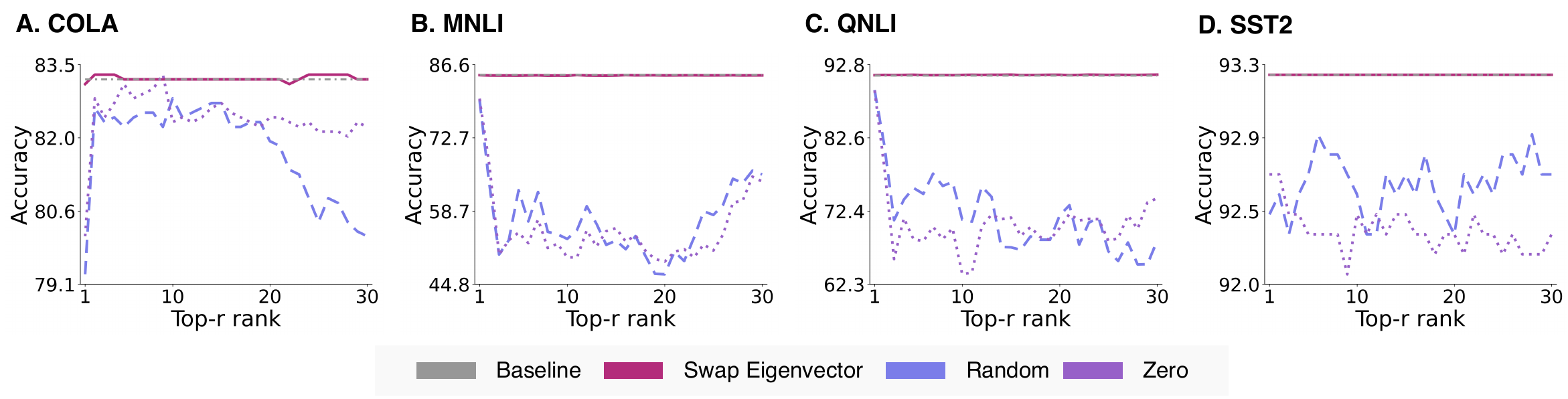}
  \caption{
    \textbf{Pretrained-to-finetuned singular-vector replacement (FFN).}
    Replacing finetuned singular vectors with pretrained ones causes only minor performance 
    degradation across tasks, especially for leading components.
  }
  \label{fig:Appendix_finetune_pretrain_swap_FC}
\end{figure*}

\begin{figure*}[!t]
  \centering
  \includegraphics[width=0.98\linewidth]{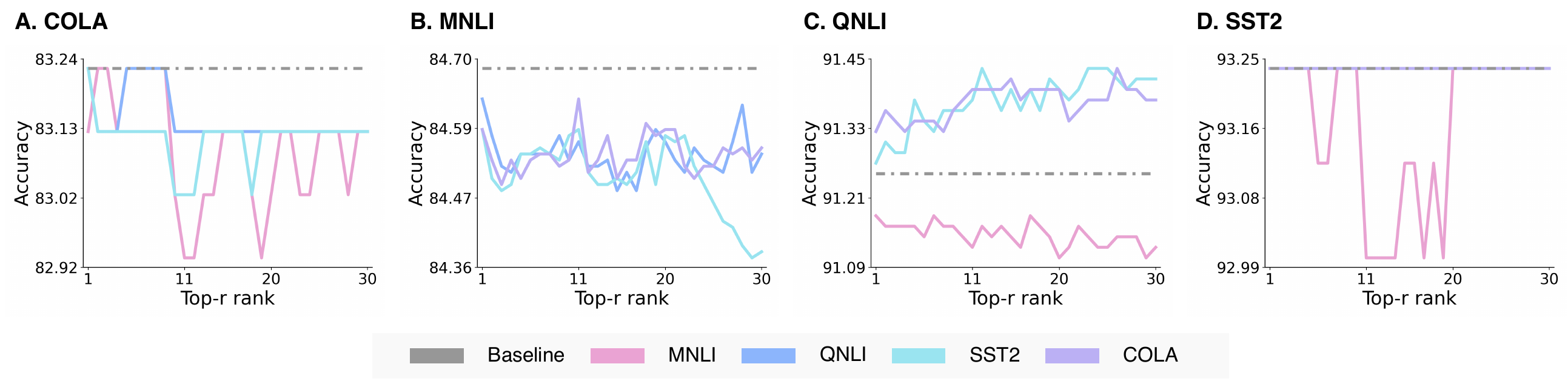}
  \caption{
    \textbf{Cross-task singular-vector swapping (FFN).}
    Swapping singular vectors between models finetuned on different tasks has limited impact 
    on performance in most cases, indicating strong cross-task compatibility in spectral directions.
      }
  \label{fig:Appendix_finetune_finetune_swap_FC}
\end{figure*}

\paragraph{Feed-forward layers.}
Figures~\ref{fig:Appendix_finetune_pretrain_swap_FC} and 
\ref{fig:Appendix_finetune_finetune_swap_FC} show the full results for FFN layers. Across all tasks, replacing finetuned singular vectors with pretrained ones results in only minor performance degradation, particularly when restricted to leading components. Similarly, cross-task swapping has limited impact in most cases. These results indicate that finetuning induces only small rotations of the spectral basis, and that different tasks remain largely compatible in the singular vectors they use.

\begin{figure*}[!t]
  \centering
  \includegraphics[width=0.98\linewidth]{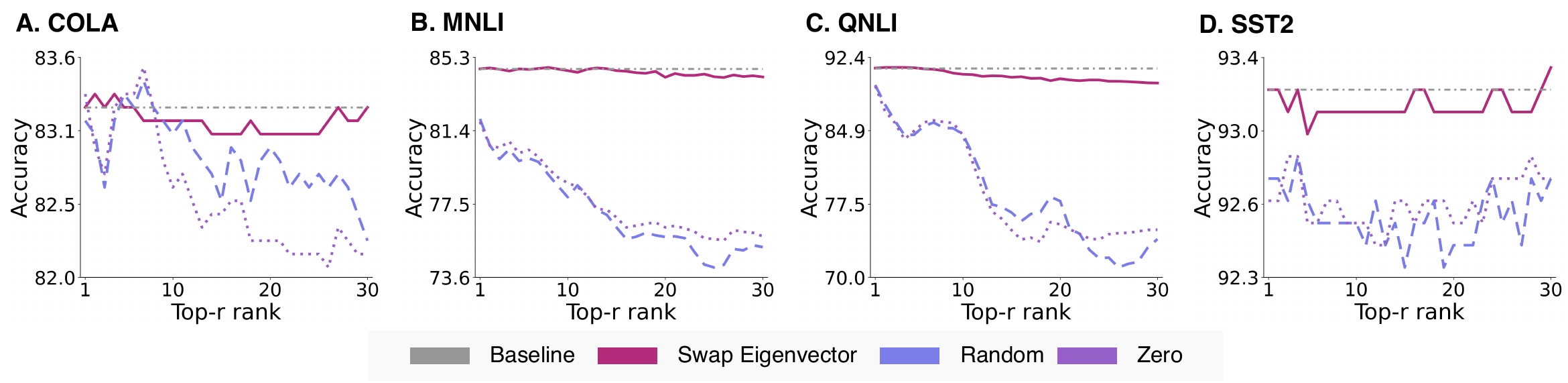}
  \caption{
    \textbf{Pretrained-to-finetuned singular-vector replacement (attention).}
    Attention projections show similar robustness: replacing singular vectors leads to only 
    modest performance changes, particularly in leading components.
  }
  \label{fig:Appendix_finetune_pretrain_swap_attention}
\end{figure*}

\begin{figure*}[!t]
  \centering
  \includegraphics[width=0.98\linewidth]{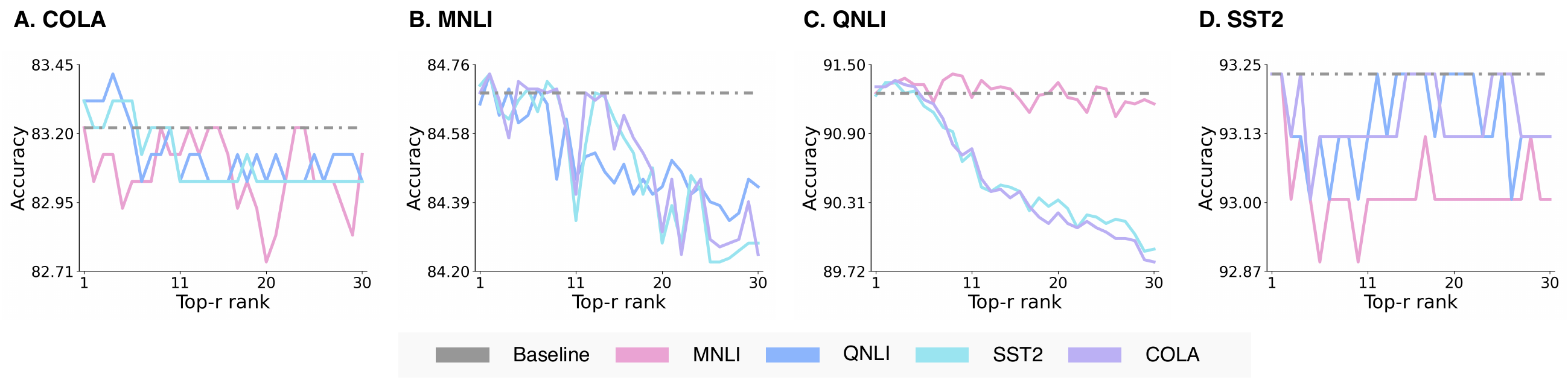}
  \caption{
    \textbf{Cross-task singular-vector swapping (attention).}
    Performance remains largely stable under cross-task swapping, suggesting that attention 
    modules also share a compatible spectral basis across tasks.
  }
  \label{fig:Appendix_finetune_finetune_swap_attention}
\end{figure*}

\paragraph{Attention modules.}
Figures~\ref{fig:Appendix_finetune_pretrain_swap_attention} and 
\ref{fig:Appendix_finetune_finetune_swap_attention} show analogous results for attention projections. The same qualitative behavior is observed: performance is largely preserved under both pretrained-to-finetuned replacement and cross-task swapping, especially in the top-ranked spectral components. While sensitivity varies somewhat across tasks and ranks, the overall conclusion remains consistent with FFN layers.

Across both module types and all tasks, swapping singular vectors leads to only limited performance changes. This supports a unified view: finetuning primarily operates by reweighting spectral coefficients within a stable pretrained basis, rather than learning substantially new singular vectors.

\section{Information Content Captured by Top-K Singular Vectors}
\label{appendix:singular_value_variance}





In the main text, we primarily analyzed the top-30 singular vectors of parameter matrices, and in Appendix~\ref{appendix:MoreRank} we extended this analysis to the top-200 singular vectors. Here we provide a complementary analysis of how much variance these leading spectral components explain across different models and ranks, to contextualize the scope and implications of our findings.

\begin{figure*}[htbp]
  \centering
  \includegraphics[width=0.9\linewidth]{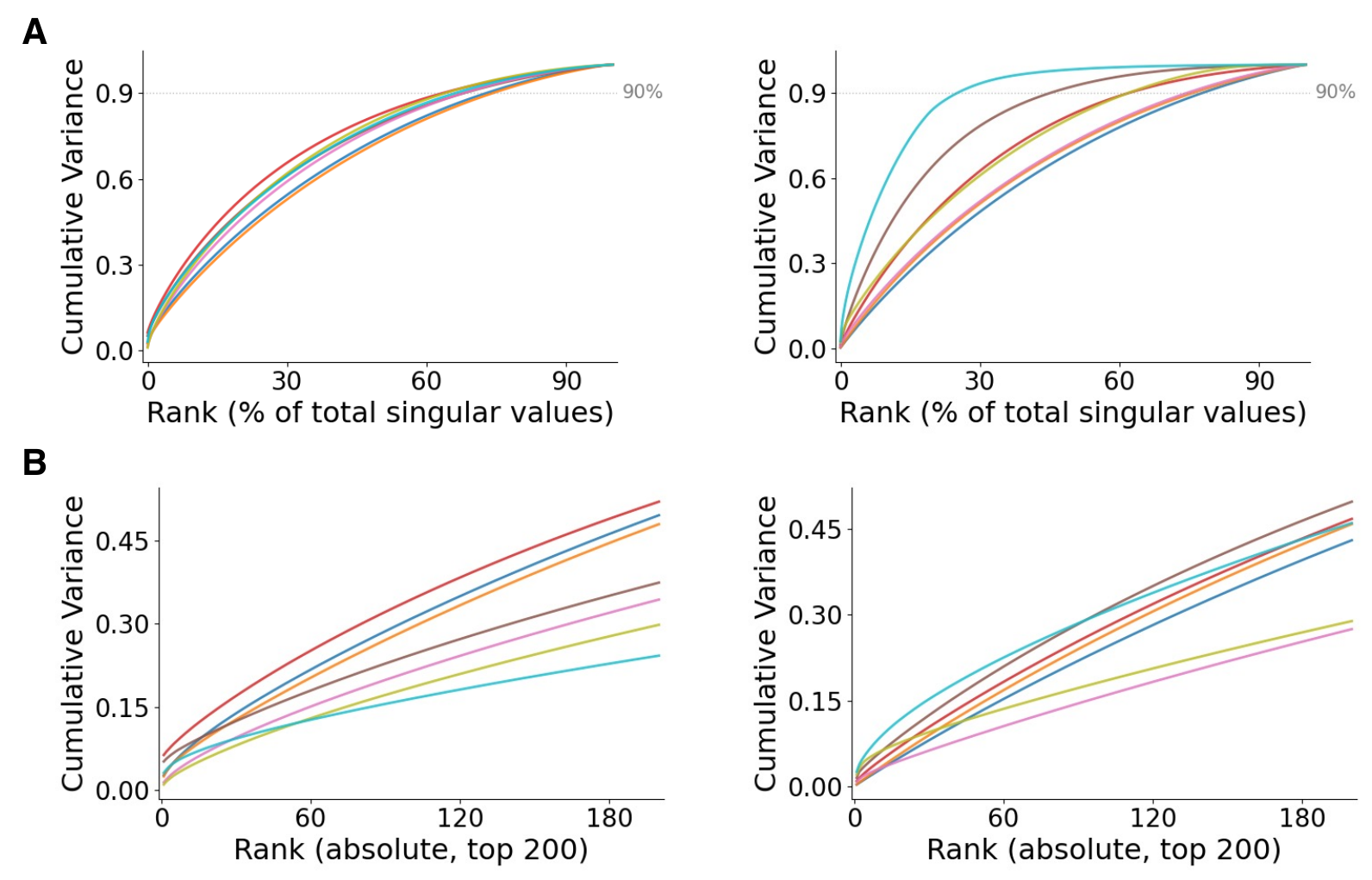}
    \caption{
    \textbf{Cumulative explained variance of ConvNeXt last-block FC weights.}
    The first and second columns correspond to \textbf{fc1} and \textbf{fc2}, respectively.
    \textbf{(A)} Explained variance versus relative rank. About 60\% of singular values are needed to capture 90\% of the Frobenius energy, with fc1 spectra slightly more concentrated than fc2.
    \textbf{(B)} Explained variance of the top singular values in absolute rank. The top-30 components explain about 10--15\% of the variance, while the top-200 explain 30--40\%. PEFT and replacement results indicate that task-relevant information is concentrated in the leading spectral components.
    }
  \label{fig:Appendix_SingularValue_variance}
\end{figure*}

Figure~\ref{fig:Appendix_SingularValue_variance} shows the cumulative explained variance as a function of rank for the last-block MLP weights (fc1 and fc2) across ConvNeXt models of varying scales. Panel (A) presents results as a function of relative rank (percentage of total singular values), while Panel (B) shows the absolute rank up to the first 200 components. Several patterns emerge. First, when measured in relative terms, approximately 60\% of the total rank is required to capture 90\% of the energy, with fc1 layers (first column) exhibiting slightly more concentrated spectra than fc2 (second column). Second, in absolute terms, the top-30 singular vectors account for roughly 10--15\% of the total variance, while the top-200 components explain approximately 30--40\%, depending on the model.

At first glance, the relatively low explained variance of the top-30 or even top-200 singular vectors might suggest that our analysis captures only a limited view of the full parameter space. However, two key pieces of evidence demonstrate that the leading spectral directions disproportionately determine downstream task performance, even though they represent a minority of the total spectral energy. First, as shown in Table \ref{tab:peft_baselines} and Section \ref{appendix:MoreRank}, finetuning only the top-300 singular components, which explain less than 50\% of total variance in most layers, achieves performance comparable to full finetuning on GLUE benchmarks. This result indicates that the bulk of the spectral mass in the tail contributes little to task-specific adaptation. The low-variance directions, despite accounting for the majority of the parameter count, appear to play a negligible role in determining model behavior on downstream tasks.

Second, our singular vector masking experiments in Section \ref{appendix:mask order} provide direct evidence that the lower half of the spectrum is functionally inert. When we progressively mask singular vectors starting from the tail (lowest-ranked components), model performance remains virtually unchanged until a substantial portion is removed. In contrast, masking from the head (highest-ranked components) causes immediate performance degradation. This striking asymmetry reveals that although the tail singular vectors encode substantial information about the parameter matrix, this information contributes little to model generalization. The leading directions concentrate the task-relevant structure, while the tail of the spectrum serves primarily as a reservoir of unused representational capacity.

Taken together, these observations indicate that while pretrained weight matrices possess a long tail of low-variance singular values, this tail plays a limited role in transfer learning. The leading singular vectors, despite accounting for a modest fraction of total variance, capture the geometry that matters for downstream adaptation and generalization. Our focus on the top-30 to top-200 components is therefore not a limitation of our analysis, but rather a deliberate emphasis on the spectral subspace where both pretraining and finetuning concentrate their representational capacity.

\section{Sensitivity of Masking Order }
\label{appendix:mask order}

To investigate the role of spectral rank ordering, we compare two complementary masking strategies: (i) masking from the top rank downward, and (ii) masking from the bottom rank upward.

\begin{figure*}[htbp]
  \centering
  \includegraphics[width=0.98\linewidth]{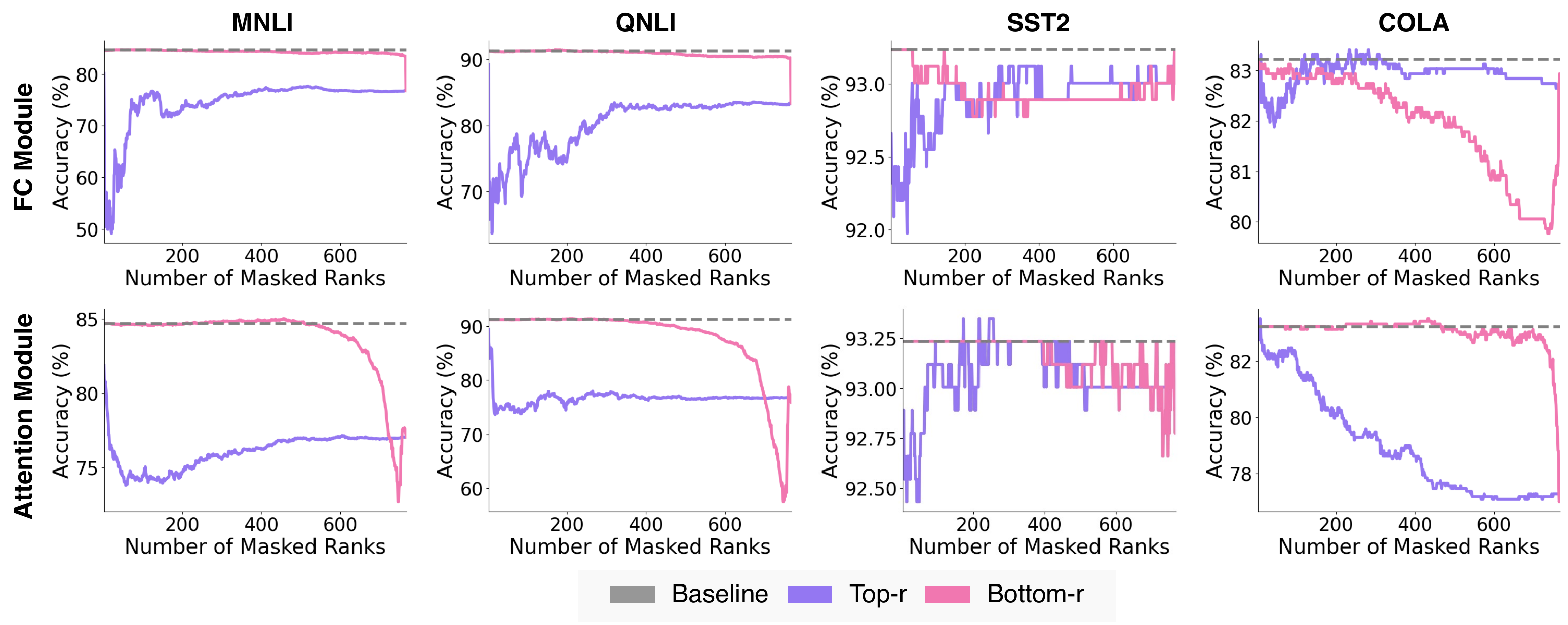}
  \caption{
    \textbf{Masking order comparison across tasks and module types.}
    Performance under top-down masking (from leading ranks) versus bottom-up masking (from trailing ranks) on four datasets. Results are shown for both feed-forward (FC) and attention modules.
  }
  \label{fig:Appendix_mask_order}
\end{figure*}

Figure~\ref{fig:Appendix_mask_order} shows that these two strategies yield markedly different behaviors. When masking begins from the top ranks, performance degrades rapidly, indicating that leading singular components carry critical task-relevant information. In contrast, masking from the bottom ranks preserves performance remarkably well, in some tasks, over 90\% of the trailing singular vectors can be masked with negligible impact. 

This asymmetry reveals a key insight: although the top-$K$ singular vectors account for a relatively small fraction of the total spectral energy (as shown in Appendix~\ref{appendix:singular_value_variance}), they disproportionately determine downstream task performance. While these leading components may explain less than 50\% of the parameter variance, they capture the directions most relevant for generalization. Conversely, the majority of the spectral basis, despite representing most of the parameter mass, remains largely task-agnostic and can be safely discarded without affecting performance. This reinforces our central finding that task-specific information is concentrated in a small number of leading spectral components, and that the pretrained spectral basis provides a reusable coordinate system where only a few directions require adaptation.

\section{Relation between Singular Vector Similarity and Generalization Performance}
\label{appendix:generalization}

\begin{figure*}[!t]
  \centering
  \includegraphics[width=0.98\linewidth]{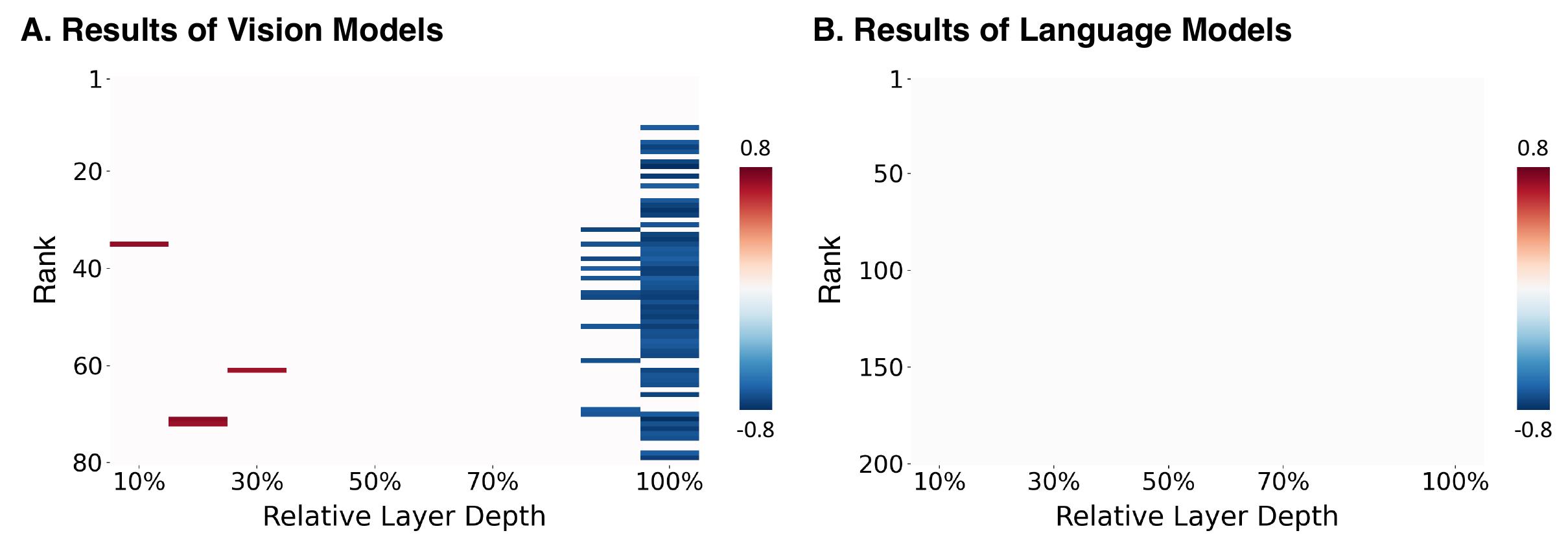}
  \caption{\textbf{Singular vector similarity versus generalization performance across layers and rank ranges.} 
    \textbf{(A)} Vision models. 
    \textbf{(B)} Language models. 
    Across both modalities, no consistent or statistically significant relationship between singular vector similarity and generalization performance is observed after multiple testing correction.
}
  \label{fig:Appendix_generalization}
\end{figure*}

We investigate the relationship between singular vector similarity (before and after finetuning) and generalization performance, with statistical significance assessed under false discovery rate (FDR) correction. Figure~\ref{fig:Appendix_generalization} summarizes the results across layers, modalities, and rank ranges.

After controlling for multiple comparisons, we do not observe any statistically significant or consistent correlation between singular vector similarity and generalization performance. This suggests that singular vector similarity, by itself, is not a reliable or robust predictor of downstream generalization.

We attribute this lack of consistent relationship primarily to heterogeneity in finetuning strategies across models. In practice, different models adopt substantially different optimization regimes for pretrained components. For example, some configurations effectively freeze the encoder or apply very small learning rates, resulting in minimal changes to singular vectors and thus high similarity, regardless of downstream performance. Consequently, high similarity can arise both when a model generalizes well and when it fails to adapt, weakening any direct association between similarity and performance.

\section{Additional Details for the Cumulative-Gradient Analysis}
\label{appendix:Gradient}




In the main text, Figure \ref{fig:gradient} demonstrates that finetuning updates are strongly confined to the spectral basis of pretrained models through top singular vector analysis. Here we extend this analysis in two directions: (1) examining more rank levels to understand how spectral alignment varies across different ranks, and (2) providing results across a broader collection of pretrained architectures to validate the generality of our findings.

\begin{figure*}[!t]
  \centering
  \includegraphics[width=0.98\linewidth]{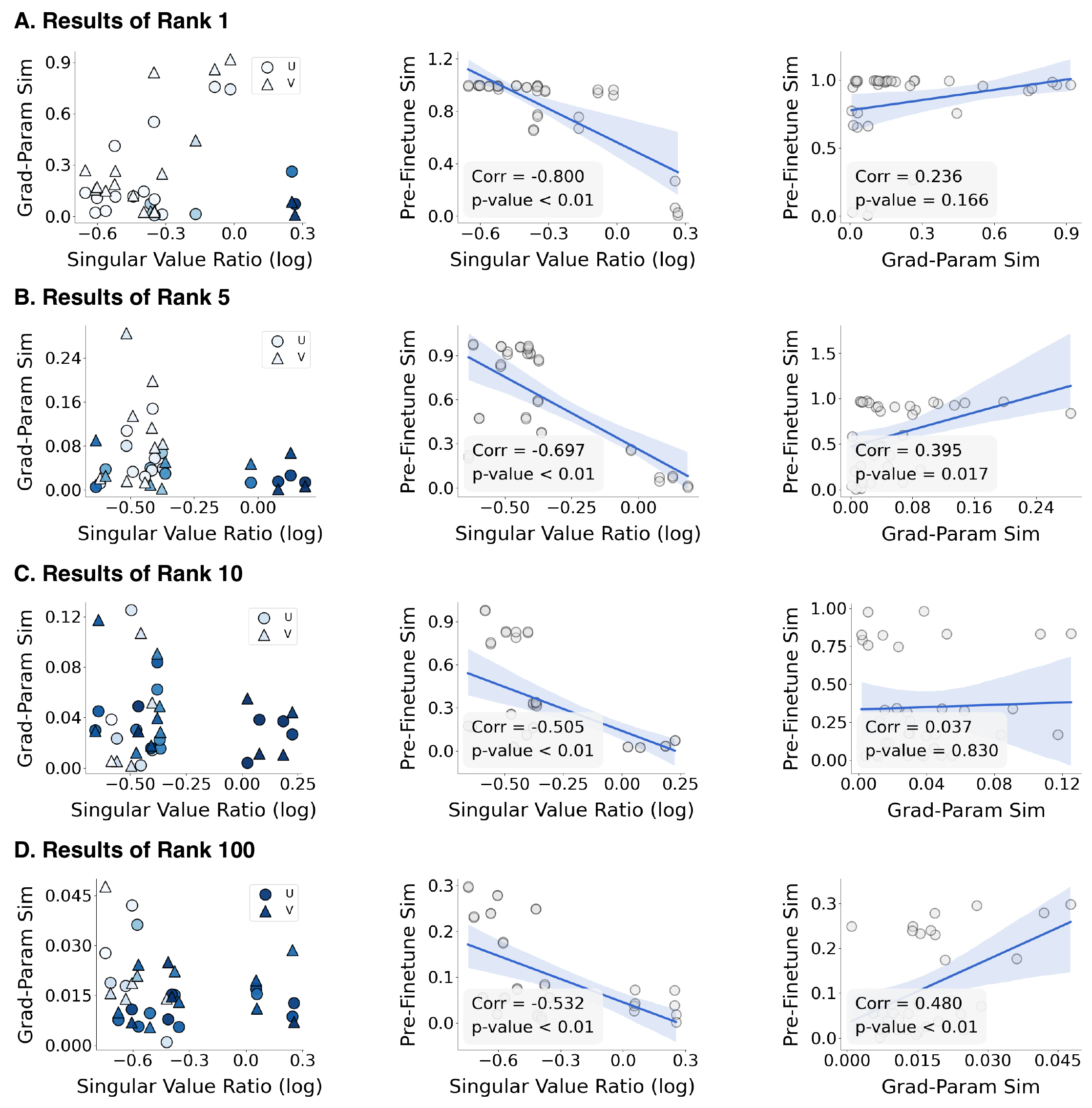}
  \caption{
  \textbf{Multi-rank spectral alignment analysis.}
  Smaller singular value ratios correlate with higher vector similarity. Alignment decreases significantly at higher ranks, with rank-100 showing maximum similarity below 0.1.
  }
  \label{fig:Appendix_gradient_rank}
\end{figure*}

Figure~\ref{fig:Appendix_gradient_rank} presents the cumulative-gradient analysis across rank-1, rank-5, rank-10, and rank-100 singular vectors. Several consistent patterns emerge across all rank levels. First, we observe a clear inverse relationship: the smaller the ratio between cumulative gradient singular values and pretrained parameter singular values, the higher the cosine similarity between their corresponding singular vectors before and after finetuning. This indicates that when finetuning introduces relatively small perturbations in the parameter, it tends to preserve the pretrained spectral directions more faithfully.

Second, we find that the similarity between cumulative gradient and pretrained parameter singular vectors also correlates with the preservation of singular vector directions, though this relationship is less pronounced. 

More importantly, we observe that as rank increases, both types of similarity decline significantly. At rank-100, the maximum similarity drops below 0.1, revealing that the coupling between gradients and pretrained parameters is most evident in the leading singular vectors. This suggests that finetuning primarily operates within a low-dimensional subspace spanned by the leading spectral components of the pretrained model.

\begin{figure*}[!t]
  \centering
  \includegraphics[width=0.98\linewidth]{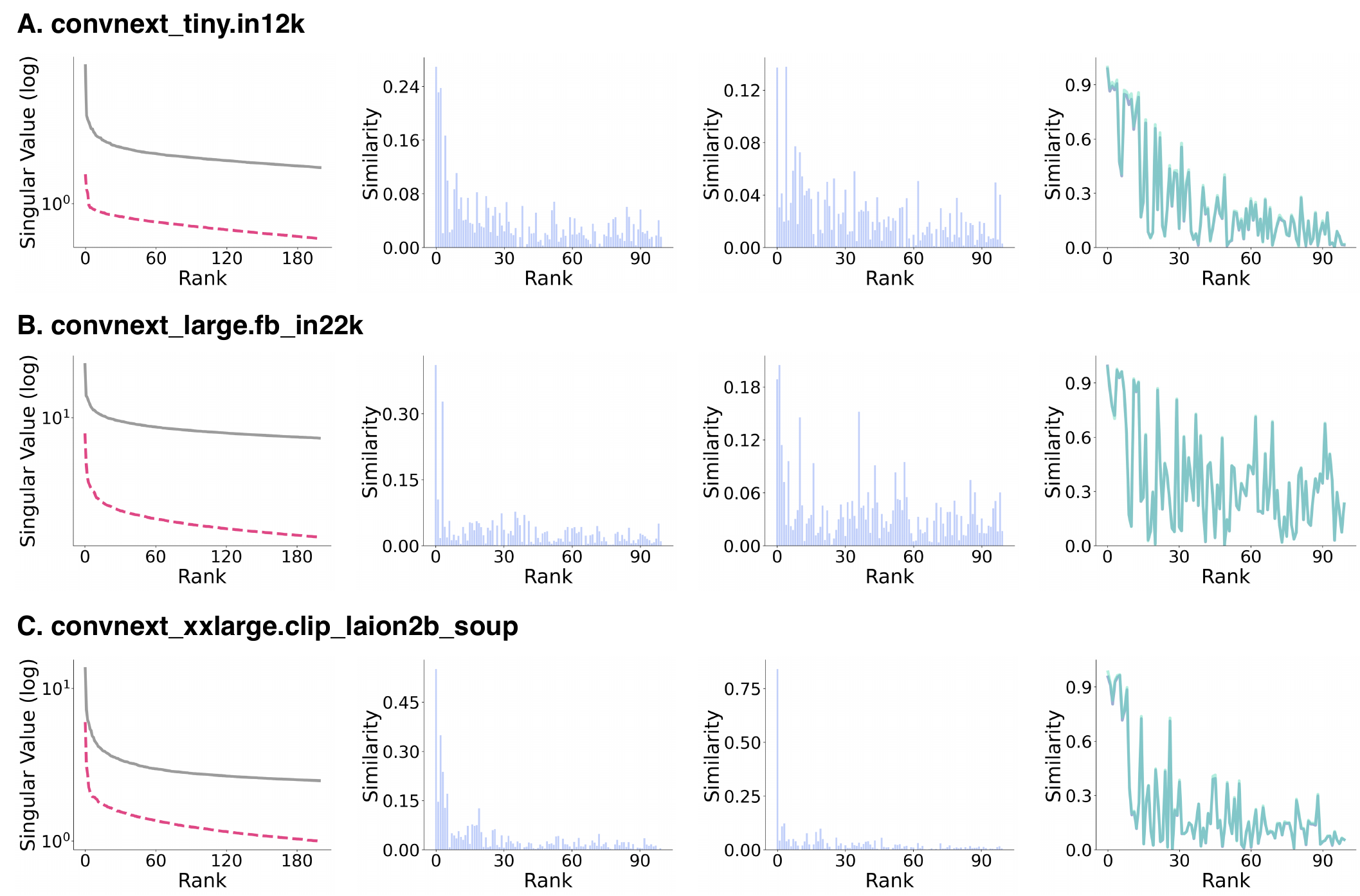}
  \caption{
  \textbf{Cumulative-gradient analysis across pretrained models.}
  Gradient singular values are consistently smaller than pretrained parameter singular values across all models, indicating small-magnitude finetuning perturbations.
  }
  \label{fig:Appendix_gradient}
\end{figure*}

Figure~\ref{fig:Appendix_gradient} presents results for several models across different architectures and scales. Across all examined models, the singular values of cumulative gradients are substantially smaller than those of the corresponding pretrained weight matrices. This indicates that the magnitude of updates during finetuning is small relative to the pretrained parameter scale. Consequently, the stability of leading singular vectors under finetuning is primarily attributable to the limited update magnitude: since parameter changes are small, the dominant spectral structure of pretrained weights remains largely intact throughout adaptation.

Overall, these extended results reinforce the central finding: pretraining defines a dominant spectral subspace that both stabilizes leading singular vectors and confines finetuning updates to a low-dimensional manifold, rather than allowing free exploration of the full parameter space.

\section{Additional Results on the Effect of Pretraining Data Scale on Spectral Stability}
\label{appendix:pretrain data}

\paragraph{Pretraining setup.} All pretraining experiments are conducted on CIFAR-10. We train Masked Autoencoder (MAE) models from scratch on varying amounts of training data, while fixing the validation set to the last 5,000 samples of the CIFAR-10 test set. The best checkpoint is selected based on validation loss. Images are resized to 32*32 and divided into 4*4 patches, yielding 64 patches per image. We use a masking ratio of 60\%. The encoder follows a ViT-Tiny configuration: embedding dimension 192, 6 Transformer blocks, 3 attention heads, and 2D sinusoidal positional encoding. The decoder is lightweight: embedding dimension 64, 2 Transformer blocks, 4 attention heads, followed by a linear projection to predict pixel values for each patch (dimension 4*4*3=48). The reconstruction objective is per-patch MSE loss. We train for 200 epochs with batch size 64 using AdamW (learning rate 1.5e-3, weight decay 1e-4), cosine annealing learning rate schedule, gradient clipping at 1.0, and mixed precision training (AMP + GradScaler) with TF32 acceleration enabled. Validation is performed every epoch, and weights are saved when validation loss reaches a new minimum.

\paragraph{Post-training scenarios.} We conduct two types of post-training experiments starting from the best pretrained checkpoints at each data scale. For same-task continued training, we further train the MAE model on the first 1,000 samples of the CIFAR-10 test set, maintaining the same architecture and reconstruction objective as pretraining. We train for 10 epochs with batch size 64 using AdamW (learning rate 3e-3, weight decay 1e-4), cosine annealing schedule, and gradient clipping at 1.0. For classification finetuning, we reuse the pretrained encoder and adding a classification head. Specifically, we perform global average pooling over all patch tokens from the encoder output, followed by a linear layer. Only encoder parameters are loaded from pretrained checkpoints, while the classification head is randomly initialized. We finetune on the first 3,000 samples of the CIFAR-10 test set for 10 epochs with batch size 64, using cross-entropy loss, AdamW optimizer (learning rate 1.5e-3, weight decay 1e-4), cosine annealing schedule, and gradient clipping at 1.0.

In the main text, we illustrate the post-training evolution of singular-vector similarity for two representative pretraining data scales, 1{,}000 and 50{,}000 samples, to show that larger pretraining datasets lead to a more stable and transferable spectral basis. Here we provide the complete results across a broader range of data scales.

\begin{figure*}[!t]
  \centering
  \includegraphics[width=0.98\linewidth]{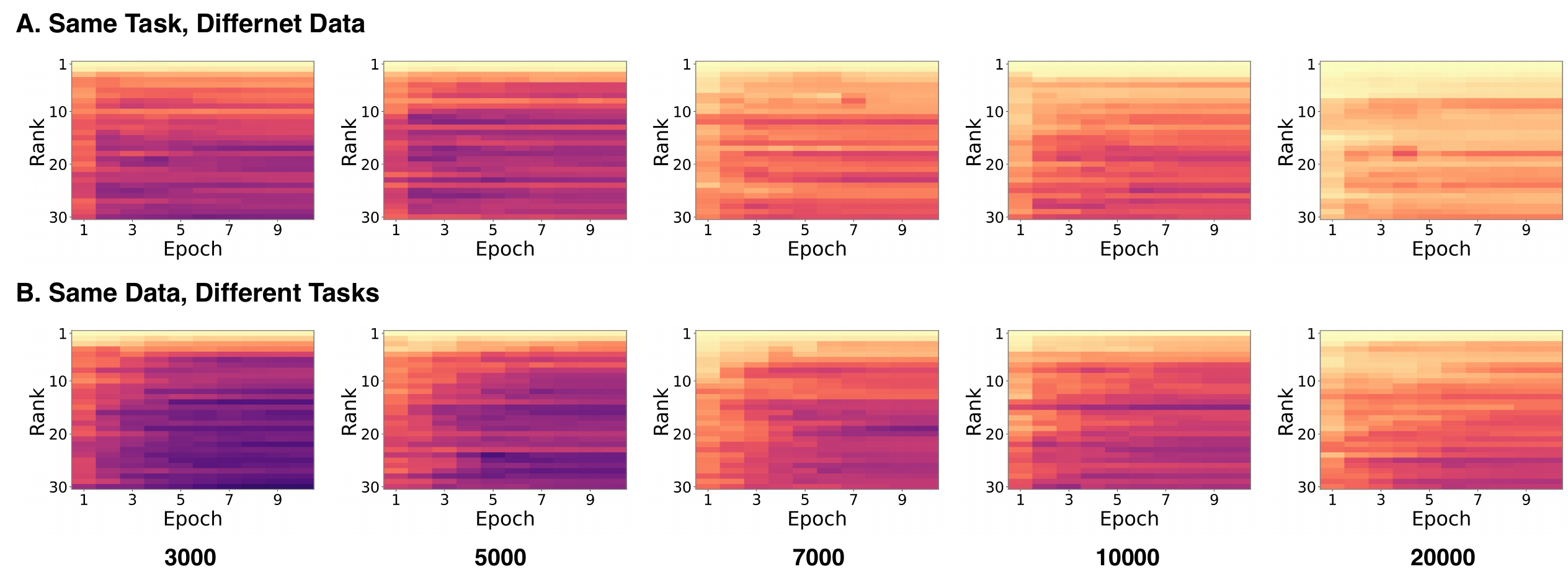}
  \caption{
  \textbf{Additional results across more pretraining data scales.}
    \textbf{(A)} Post-training on the same task with different data.
    \textbf{(B)} Post-training on the same data source with different tasks.
    In both settings, models pretrained on larger datasets exhibit more stable singular vectors during post-training.
  }
  \label{fig:Appendix_pretrain_data_amount}
\end{figure*}

Figure~\ref{fig:Appendix_pretrain_data_amount} shows the singular-vector similarity between the current model and its pretrained initialization throughout post-training, for models pretrained with different amounts of data but using the same architecture and training procedure. Across all settings, similarity decreases as adaptation proceeds, indicating that post-training progressively rotates the pretrained spectral basis. However, the rate and extent of this rotation depend strongly on pretraining data scale: models pretrained on larger datasets consistently retain higher similarity during training and converge to more aligned spectral bases.

This pattern holds not only for the extreme cases shown in the main text, but also for intermediate data scales. The transition is gradual and monotonic overall, suggesting that spectral transferability improves continuously with pretraining data scale rather than emerging only at very large dataset sizes. These results further support our main claim that larger-scale pretraining produces singular vectors that are more reusable under subsequent task or distribution shifts.

\section{Implementation Details for Spectral Parameter-Efficient Transfer}
\label{appendix:peft_details}

This section provides implementation details for the spectrally restricted parameter-efficient finetuning experiments in Section~\ref{subsec:peft}. 
The goal of these experiments is to test whether downstream adaptation can be performed effectively by updating only a small number of coordinates in the singular-vector basis of the pretrained model.

\paragraph{Base model and task setup.}
All experiments use \texttt{microsoft/deberta-v3-base} as the pretrained backbone. 
For the controlled ablation experiments in Figure~\ref{fig:peft}, we use MNLI as the representative downstream task. 
MNLI is a three-way natural language inference task with labels \texttt{entailment}, \texttt{contradiction}, and \texttt{neutral}. 
The maximum sequence length is set to $256$. 
Unless otherwise specified, the random seed is fixed to $42$.

\paragraph{Spectral parameterization.}
We apply the spectrally restricted adapter to the feed-forward module weights of DeBERTa. 
For each selected pretrained weight matrix, we compute its singular value decomposition
$W_0 = U \Sigma V^\top .$
The pretrained singular vectors $U$ and $V$ are kept fixed throughout finetuning, and only a small block of spectral coefficients is made trainable. 
Specifically, we select a contiguous spectral interval $[k,m)$, where $k$ is the starting singular-coordinate index and $m$ is the ending index. 
The effective trainable rank is therefore
$r = m-k .$
For example, $k=0$ and $m=300$ means that the method adapts the first $300$ pretrained spectral coordinates. 
This restricts finetuning to a low-dimensional subspace defined by the pretrained spectral basis.

For DeBERTa-v3-base, the relevant FFN dimension is $d=768$, so the valid spectral index range is bounded by $0 \leq k < m \leq 768$. 
In the rank-sweep experiments, we vary $m$ in increments of $50$.

\paragraph{Target layers and trainable parameters.}
By default, the spectral adapter is applied only to the last two DeBERTa encoder blocks, i.e., blocks $10$ and $11$. 
Within each selected block, we modify the FFN matrices \texttt{intermediate.dense} and \texttt{output.dense}. 
All other pretrained backbone parameters are frozen.

\paragraph{Ablation experiments in Figure~\ref{fig:peft}.}
Figure~\ref{fig:peft} contains three controlled experiments on MNLI. 
These experiments are designed to evaluate the effects of spectral rank, layer depth, and the location of the selected spectral block.

\begin{itemize}[leftmargin=0pt]
    \item \textbf{Experiment 1: rank-budget sweep.}
    We fix $k=0$ and vary
    $m \in \{50,100,150,\ldots,700\}.$
    Since $k=0$, the effective trainable rank is $r=m$. 
    This experiment tests whether increasing the number of rank improves downstream performance.

    \item \textbf{Experiment 2: layer-location comparison.}
    We fix the spectral range to $k=0$ and $m=200$, and vary the layers to which the adapter is applied. 
    This experiment tests whether adapting deeper layers leads to better generalization performance.

    \item \textbf{Experiment 3: spectral-block comparison.}
    We fix the block width to $m-k=200$ and vary the starting index $k$, e.g.,
    $k \in \{0,50,100,\ldots\}.$
    The ending index $m$ is adjusted accordingly to keep the rank budget fixed. 
    This experiment tests how adapting different spectral subspaces affects generalization performance under a fixed rank budget.
\end{itemize}

For these MNLI ablations, we train for $3$ epochs with batch size $16$, learning rate $1	\times 10^{-4}$, weight decay $0.0$, and warmup ratio $0.06$. 
We use the full training set. 
The default target blocks are $[10,11]$ unless the layer-location experiment specifies otherwise.

\paragraph{GLUE benchmark setting.}
For the full GLUE comparison in Table~\ref{tab:peft_baselines}, we use the same DeBERTa-v3-base backbone and apply the spectral adapter to the final two FFN blocks. 
For all GLUE tasks, we adapt the leading $300$ pretrained spectral coordinates in the FFN matrices of layers $10$ and $11$. 
The batch size is fixed to $16$, weight decay is set to $0.0$, the warmup ratio is $0.06$, and the random seed is $42$ for all tasks. The maximum sequence length is $256$.

Table~\ref{tab:peft_glue_task_hparams} reports the task-specific hyperparameters, namely the number of training epochs and the learning rate.

\begin{table}[htbp]
\centering
\caption{
Task-specific hyperparameters for the GLUE experiments in Table~\ref{tab:peft_baselines}. 
}
\label{tab:peft_glue_task_hparams}
\begin{tabular}{lcc}
\toprule
\textbf{Task} & \textbf{Epochs} & \textbf{Learning Rate} \\
\midrule
SST-2 & 20 & $1 \times 10^{-4}$ \\
MRPC  & 12 & $2 \times 10^{-4}$ \\
MNLI  &  5 & $2 \times 10^{-4}$ \\
CoLA  & 30 & $3 \times 10^{-4}$ \\
QNLI  & 10 & $2 \times 10^{-4}$ \\
QQP   & 10 & $2 \times 10^{-4}$ \\
RTE   & 20 & $2 \times 10^{-4}$ \\
STS-B & 10 & $2 \times 10^{-4}$ \\
\bottomrule
\end{tabular}
\end{table}

\paragraph{Optimization and implementation details.}
The SVD of each selected pretrained weight matrix is computed once before downstream training. 
The singular vectors are then frozen, and finetuning optimizes only the selected spectral coefficients and the task-specific prediction head. 
We use the AdamW optimizer with a linear learning-rate schedule and warmup, where the warmup phase occupies $6\%$ of the total training steps. No weight decay is used.

For robustness when downloading pretrained checkpoints or datasets, we allow up to $5$ retries. 
The base waiting time between retries is $8$ seconds. 
These retry parameters only affect data and checkpoint loading and do not change the training procedure.

\paragraph{Computational resources.}
All experiments are conducted on a single NVIDIA GeForce RTX 4090 GPU with 24GB of memory. The CPU is an AMD EPYC 7T83 64-Core Processor.

\paragraph{Summary.}
These implementation choices make the PEFT experiment a direct test of the paper's central hypothesis. 
The pretrained singular-vector basis is fixed, and downstream learning is restricted to a small number of spectral coordinates in the final FFN layers. 
The competitive GLUE performance obtained under this constraint indicates that a compact subset of pretrained spectral directions already provides an effective basis for parameter-efficient transfer.



\end{document}